\documentclass[sigconf]{acmart}
\usepackage{epstopdf}
\usepackage{multirow}
\usepackage{amsfonts}
\usepackage{bbm}
\usepackage[misc]{ifsym}

\AtBeginDocument{%
  \providecommand\BibTeX{{%
    \normalfont B\kern-0.5em{\scshape i\kern-0.25em b}\kern-0.8em\TeX}}}

\copyrightyear{2020}
\acmYear{2020}
\setcopyright{acmcopyright}\acmConference[MM '20]{Proceedings of the 28th ACM
International Conference on Multimedia}{October 12--16, 2020}{Seattle, WA, USA}
\acmBooktitle{Proceedings of the 28th ACM International Conference on Multimedia
(MM '20), October 12--16, 2020, Seattle, WA, USA}
\acmPrice{15.00}
\acmDOI{10.1145/3394171.3413692}
\acmISBN{978-1-4503-7988-5/20/10}

\begin{document}

\fancyhead{}
\title{Hard Negative Samples Emphasis Tracker without Anchors}

\author{Zhongzhou Zhang}

\affiliation{%
  \institution{School of Microelectronics and Communication Engineering, Chongqing University, China}
}
\email{zz.zhang@cqu.edu.cn}
\orcid{1234-5678-9012}

\author{Lei Zhang}

\authornote{Corresponding author.}
\affiliation{
    \institution{School of Microelectronics and Communication Engineering, Chongqing University, China}}
\email{leizhang@cqu.edu.cn}
\begin{abstract}
  Trackers based on Siamese network have shown tremendous success, because of their balance between accuracy and speed. Nevertheless, with tracking scenarios becoming more and more sophisticated, most existing Siamese-based approaches ignore the addressing of the problem that distinguishes the tracking target from hard negative samples in the tracking phase. The features learned by these networks lack of discrimination, which significantly weakens the robustness of Siamese-based trackers and leads to suboptimal performance. To address this issue, we propose a simple yet efficient hard negative samples emphasis method, which constrains Siamese network to learn features that are aware of hard negative samples and enhance the discrimination of embedding features. Through a distance constraint, we force to shorten the distance between exemplar vector and positive vectors, meanwhile, enlarge the distance between exemplar vector and hard negative vectors. Furthermore, we explore a novel anchor-free tracking framework in a per-pixel prediction fashion, which can significantly reduce the number of hyper-parameters and simplify the tracking process by taking full advantage of the representation of convolutional neural network. Extensive experiments on six standard benchmark datasets demonstrate that the proposed method can perform favorable results against state-of-the-art approaches.
\end{abstract}



\maketitle

\section{Introduction}
Visual object tracking is one of the fundamental computer vision problems, which aims to estimate the trajectory of an arbitrary visual target when only an initial state of the target is available. Generic visual tracking is a sought-after yet challenging research topic with a wide range of applications, such as self-driving cars, surveillance, augmented reality, unmanned aerial vehicles, to name a few \cite{7001050, 6619156}. Although much progress has been achieved in recent years, it remains a challenge to design robust trackers due to many factors including occlusion, out-of-view, deformation, background cluster, motion blur and so on.
\begin{figure}[t]
\centering
\includegraphics[scale=0.35]{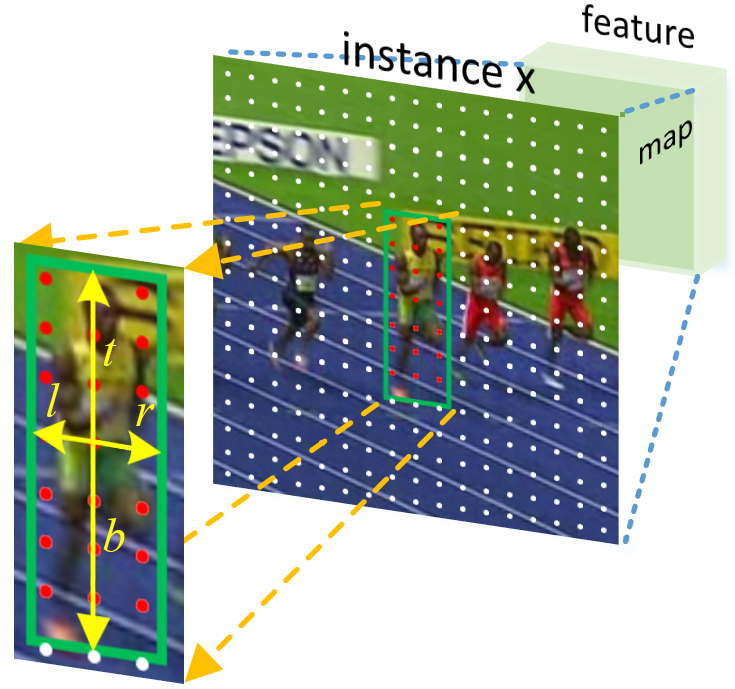}
\caption{Instantiation of pre-pixel prediction fashion. We map the points in the feature map back to the instance x. Points in red inside ground-truth (the green box) represent positive samples and points in white denote negative samples. Narrows in yellow denote the regression targets $(l^*, t^*, r^*, b^*)$.}
\label{1}
\end{figure}

Recently, Siamese-based trackers \cite{zhang2019deeper, li2018high, li2019siamrpn++, wang2019spm, fan2019siamese} have drawn great attention in visual tracking community by achieving favorable performance and high efficiency. One representative example is SiamFC \cite{bertinetto2016fully}, which learns similarity knowledge of the same object in different frames in an end-to-end off-line training method. SiamFC \cite{bertinetto2016fully} simply adopts multi-scale search to estimate the target scale. To get bounding box with different aspect ratios, Siamese-RPN \cite{li2018high} integrates Siamese subnetwork and region proposal subnetwork(RPN) \cite{ren2015faster}. By using pre-defined anchor boxes with different scales and aspect ratios, Siamese-RPN \cite{li2018high} can predict more accurate bounding boxes.
Although they have obtained good performance, there are still some problems that should be addressed.
\textbf{(1)} Anchor-based trackers bring many additional hyper-parameters, such as the number of pre-defined anchors. For a tracking task, speed is also an important indicator in practical applications. Too many pre-defined anchors severely reduce computational efficiency. \textbf{(2)} Most Siamese-based approaches ignore to address the problem of distinguishing positive samples from hard negative samples in the tracking phase. Actually, all samples of the search images can be categorized into three types: \emph{positive samples}, \emph{easy negative samples} and \emph{hard negative samples}. Positive samples are samples that near the center of the tracking target. Easy negative samples include non-semantic background and samples that are not similar to the tracking target. Compared with those easy negative samples, hard negative samples means those samples belonging to the same class of the tracking target, which are similar to positive samples in some tracking sequences. Because there is no effective means to eliminate the effect of hard negative samples, it is difficult for most Siamese-based trackers \cite{zhang2019deeper, li2018high, li2019siamrpn++, wang2019spm, fan2019siamese, bertinetto2016fully} to distinguish positive samples from hard negative samples during tracking.
Furthermore, the number of negative samples is much more than that of positive samples, so the imbalance from positive samples and negative samples also hinder better performance of Siamese-based approaches.

To solve the aforementioned problems, we first put forward an anchor-free tracking method with a per-pixel prediction fashion and the architecture is carefully designed for an end-to-end training manner. Both classification branch and regression branch in FCOS \cite{tian2019fcos} head are designed for foreground-background classification and bounding box prediction respectively.
Different from anchor-based methods, the regression target of our tracker is 4D vectors that denote the distances from points within ground-truth to the four sides of the ground-truth bounding box. As illustrated in Figure \ref{1}. It is capable of obtaining the coordinate of tracking target and a precise bounding box, which do not need many pre-defined anchors. In this way, our method can not only reduce the number of hyper-parameters but also improve the computational efficiency.

Besides, we propose a hard negative samples emphasis method, which can effectively constrain hard negative samples. General Siamese-based trackers localize the tracking target by finding the coordinate of the maximum value in the score map. However, hard negative samples also have large score responses, leading to tracking failures.
To handle this problem, we can select some candidates that include positive samples and hard negative samples, according to the ranking values in the score map. Non-Maximum Suppression (NMS) \cite{ren2015faster} is adopted to filter out redundant bounding boxes. These selected samples can be classified into positive samples and hard negative samples by their IoU values with the ground-truth.
Taking inspiration from person re-identification, we employ a contrastive loss to enlarge the distances between positive samples and hard negative samples. Also, an operation of random shifts of the ground-truth is introduced to keep the balance between positive samples and hard negative samples. The main contributions of this work are summarized as the following threefold:
\begin{itemize}
  \item We explore a new anchor-free tracking method that adopts pre-pixel prediction fashion to get accurate bounding boxes. And our method can reduce hyper-parameters and improve computational efficiency.
  \item We propose a novel hard negative samples emphasis method to overcome the weak distinguishability between positive samples and hard negative samples.
  \item We extensively evaluate the proposed method on six benchmark datasets including GOT-10k \cite{huang2019got}, UAV123 \cite{mueller2016benchmark}, VOT2016 \cite{Kristan2016a}, VOT2018 \cite{Kristan2018a}, VOT2019 \cite{Kristan2019a}, LaSOT \cite{fan2019lasot}. Our method performs well against state-of-the-art trackers.
\end{itemize}

\section{Related Work}
\textbf{Siamese Network for Tracking}. Siamese network trained with similarity learning strategy is first applied in face verification. With the development of deep learning, representative networks are introduced in the tracking community. SINT \cite{tao2016siamese} explores Siamese network to the tracking problem, which implements two-steam Convolution Neural Networks(CNNs) with shared parameters to learning a match function in the off-line training phase. During online tracking, it searches for the candidate's position in the search region by the exemplar that is obtained from the first frame. However, slow speed hinders the development of SINT. After that, Bertinetto et al. \cite{bertinetto2016fully} put forward fully convolutional Siamese network, namely SiamFC, which significantly improves the tracking speed with 86 FPS on a single GPU. SiamFC \cite{bertinetto2016fully} is trained by large-scale image pairs to learn a similarity function in an end-to-end manner. Inspired by Faster RCNN \cite{he2016deep}, Siamese-RPN \cite{li2018high} integrates Siamese network with region proposal networkp (RPN) to obtain more accurate bounding boxes. Siamese-RPN \cite{li2018high} performs a classification branch and a regression branch. Previous mentioned Siamese-based trackers almost take AlexNet \cite{krizhevsky2012imagenet} as their backbone for feature extraction. Directly using modern deep neural networks including VGG \cite{simonyan2014very}, ResNet \cite{he2016deep}, Inception \cite{szegedy2015going} into Siamese-based trackers brings about severe decrease of tracking performance. To deal with this issue, Li et al. \cite{li2018high} introduce an effective sampling method to alleviate the impact of padding operation in the training phase. Furthermore, a depth-wise cross-correlation operation \cite{li2019siamrpn++} is proposed to reduce the computational cost and enhance the representation of feature maps. Zhang et al. \cite{zhang2019deeper} present a detailed study of factors that affect tracking performance. Based on their conclusions, they design new deeper and wider networks that are specifically used for Siamese-based trackers.
Many Siamese-based trackers borrow modules or designs from object detection community. For example, SPM-tracker \cite{wang2019spm} designs a two-stage network, namely coarse matching stage and fine matching stage, which can achieve high localization precision. C-RPN \cite{fan2019siamese} adopts a multi-stage tracking framework with three RPNs cascaded and leverages features of different levels, which capable of getting remarkable performance.
\\
\\
\begin{figure*}[t]
\centering
\includegraphics[scale=0.81]{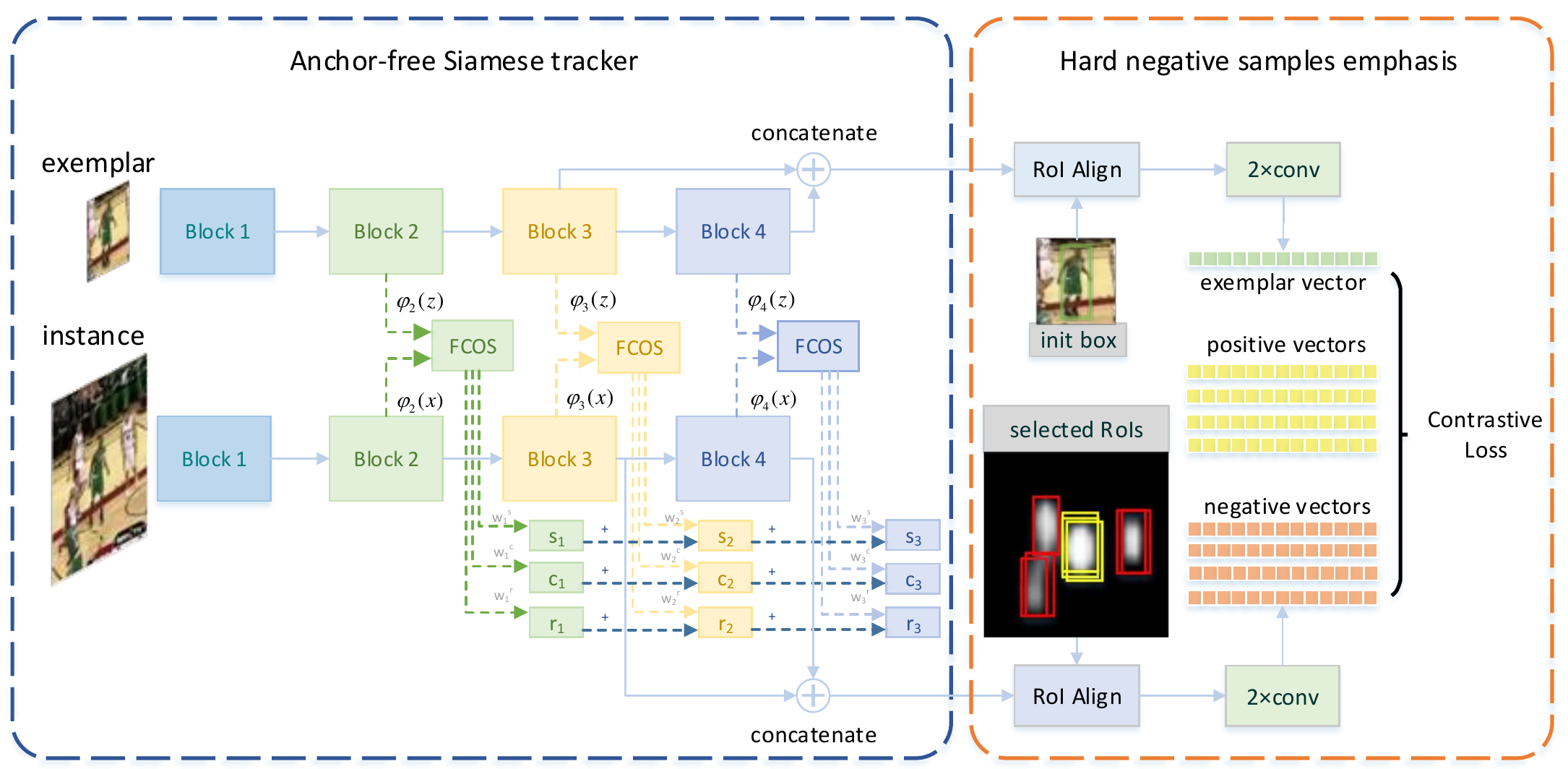}
\caption{Overview of the proposed framework. Our anchor-free Siamese tracker employs modified ResNet-50 \cite{he2016deep} as backbone for extracting multi-level features. Multi-level features are taken as the inputs of FCOS heads for score map (s), center-ness (c) and regression (r). Then RoI Align layer takes selected RoIs and concatenated features as input to obtain pooled features. Two convolutional layers are performed to make the pooled features become vectors. Finally, we adopt a contrastive loss to constrain hard negative samples.}
\label{2}
\end{figure*}
\textbf{Anchor-free Object Detection}. Object detection and object tracking have some common similarities like the RPN module mentioned earlier. Anchor-free detection methods have not been explored in the tracking community. YOLOv1 \cite{redmon2016you} is the well-known anchor-free detector, which predicts bounding boxes relying on grid cells and the grid cells are generated by input images. CornerNet \cite{law2018cornernet} introduces a new one-stage approach to object detection that detects an object as a pair of key points, the top-left corner and the bottom-right corner of bounding boxes. Recently, Tian et al. \cite{tian2019fcos} proposes a full convolutional one-stage object detection (FCOS), which gets remarkable performance on MSCOCO \cite{lin2014microsoft} benchmark. They explore a per-pixel bounding box prediction fashion. By omitting a huge number of anchors, FCOS \cite{tian2019fcos} significantly reduces hyper-parameters and avoids complicated computation.
\section{The Proposed Tracker}
We first develop an anchor-free Siamese tracker for foreground-background classification and accurate bounding box prediction in detail. The proposed tracker adopts a per-pixel prediction fashion, which takes the distances from the location of a positive sample to the four sides of the ground-truth bounding box as its regression targets. Furthermore, to tackle the problem caused by hard negative samples, we explore a hard negative samples emphasis method to maintain the discrimination of positive samples and hard negative samples in the embedding space.
\subsection{Deep Siamese Network}
In this subsection, we introduce the proposed tracking framework which is illustrated in Figure \ref{2}. We adopt the modified ResNet-50 \cite{he2016deep} as our backbone for feature extraction. The stride of original ResNet-50 is 32 pixels. To make it better suited for tracking, we adjust the effective strides of block 3 and block 4 from 16 pixels and 32 pixels to 8 pixels by the same strategy as \cite{li2019siamrpn++}. Our Siamese network consists of two branches, namely exemplar branch and instance branch, which have the same architecture and share network parameters. During training, large-scale cropped template images (exemplar z) and larger search images (instance x) are sent to exemplar branch and instance branch in pairs, extracting embedding features of exemplar z and instance x.
Due to different levels of Siamese network contain high-level and low-level features with diverse feature representations, we intend to leverage multi-level features to take full advantage of semantic information and detailed information. Multi-level features include the outputs of \emph{Block 2}, \emph{Block 3}, \emph{Block 4}. Then, we use a $1 \times 1$ convolutional layer to make sure the channels of multi-level features becoming the same. The adjusted features can be expressed as $\varphi_{i}(z)$ and $\varphi_{i}(x)$, $i \in \{ 2, 3, 4\}$.
\subsection{Anchor-free Siamese Tracker}
\label{section3.2}
\begin{figure}[t]
\centering
\includegraphics[scale=0.43]{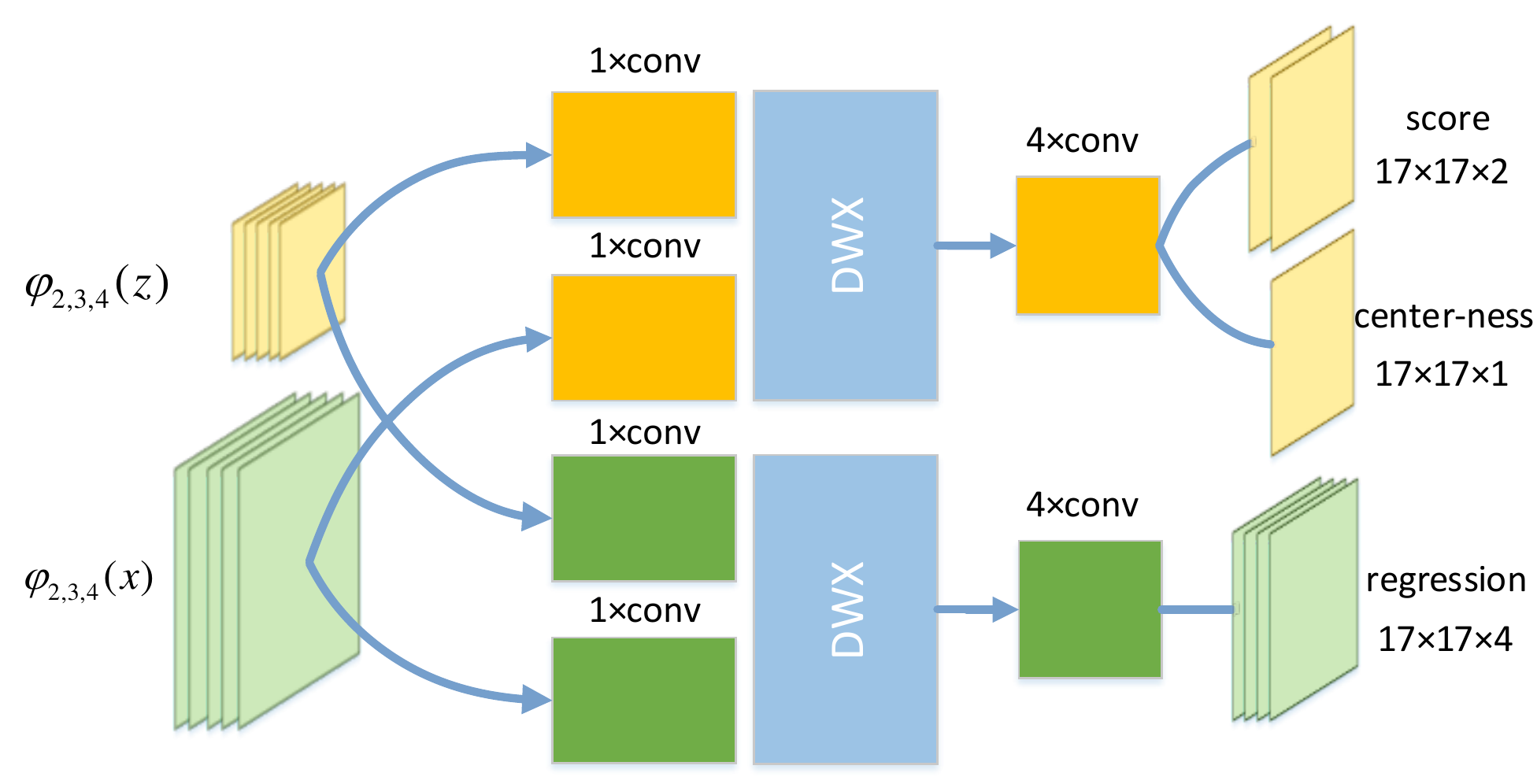}
\caption{Details of FCOS head}
\label{3}
\end{figure}

Both classification branch and regression branch in FCOS \cite{tian2019fcos} head are designed for localization and bounding box estimation.  To be specific, three FCOS heads take multi-level features of exemplar z and instance x as input. For each FCOS head, in order to make features $\varphi_{i}(z)$ and $\varphi_{i}(x)$ better suited for classification and regression tasks, we adjust them to $\{{\lbrack \varphi_i(z) \rbrack} _{cls}, {\lbrack \varphi_i(z) \rbrack} _{reg}\}$ and $\{{\lbrack \varphi_i(x) \rbrack} _{cls}, {\lbrack \varphi_i(x) \rbrack} _{reg}\}$ by a convolutional layer. Then, each branch combines the adjusted features with Depthwise Cross Correlation operation (DW-XCorr) \cite{li2019siamrpn++}, which is formulated as
\begin{equation}
\begin{split}
S_{w\times h\times 2}=\psi_s({\lbrack \varphi(z) \rbrack} _{cls} \star {\lbrack \varphi(x) \rbrack} _{cls}),\\
R_{w\times h\times 4}=\psi_r({\lbrack \varphi(z) \rbrack} _{reg} \star {\lbrack \varphi(x) \rbrack} _{reg}).
\end{split}
\end{equation}
where $S_{w\times h\times 2}$ and $R_{w\times h\times 4}$ represent the outputs of classification branch and regression branch, $w, h$ denote width and height of the output, $\psi$ represents four $3 \times 3$ convolutional layers.
In this way, we can learn a similarity match and achieve efficient information association of exemplar z and instance x in the embedding space. FCOS head is shown in Figure \ref{3}.

For the classification branch, we divide it into two sub-branches: a score map sub-branch for localization and a center-ness \cite{tian2019fcos} sub-branch for the selection of a better bounding box. To obtain the labels of the score map, we map each pixel point $(x, y)$  in the score map back to instance x according to the following relationship:
\begin{equation}
\begin{split}
 X =\lfloor \frac{s}{2} \rfloor+ sx, \quad Y = \lfloor\frac{s}{2}\rfloor + sy.
 \end{split}
\end{equation}
where $s$ represents the stride of Siamese network. If a point $(X, Y)$ in the instance x falls into the ground-truth bounding box, whose corresponding point $(x, y)$ in the score map is regarded as a positive sample and we set its label to 1. On the contrary, if the points fall outside the ground-truth bounding box, which are seen as negative samples. In short, it is a binary classification problem that distinguishes foreground from background. Figure \ref{1} illustrates the mapping process and the selection of samples. Red points and white points represent positive samples and negative samples respectively.

For the regression branch, the output is a 4 channels feature map. As mentioned earlier, we adopt a per-pixel prediction fashion that each positive sample has the potential to generate a bounding box. Each positive sample corresponds to a 4D vector $(l^*, t^*, r^*, b^*)$ that represents the distances from $(X, Y)$ to four sides of the ground-truth bounding box in the instance \textbf{x}. The training regression targets can be formulated as:
\begin{equation}
\begin{split}
l^* = X - x_0,\qquad t^* = Y - y_0, \\
r^* = X + x_1,\qquad b^* = Y - y_1.
\end{split}
\end{equation}
where $(x_0, y_0)$ and $(x_1, y_1)$ represent top-left and bottom-right coordinates of the ground-truth. Compared with the trackers based on RPN \cite{li2018high, li2019siamrpn++} that regress the target bounding boxes with pre-defined anchor boxes as reference, our anchor-free method can directly predict bounding boxes and reduce hyper-parameters. Furthermore, because our approach uses pixel-level features that hardly bring about a wrong match between the score map and the regression output, our per-pixel prediction fashion is more suitable for the tracking task.

Although all of the positive samples and their corresponding 4D vectors can generate bounding boxes of the tracking target, there are still many low-quality predicted bounding boxes that should be suppressed. Otherwise, these bad bounding boxes may cause the degradation of localization accuracy.  To select the boxes that have high IoU (Intersection of Union) with ground-truth, we add a new center-ness sub-branch that parallels with the score map sub-branch. The center-ness sub-branch is depicted as the normalized distance from the location to the center of the object. We simply implement it by a $1 \times 1$ convolutional layer. And the center-ness targets can be described as:
\begin{equation}
Centerness=\left\{
\begin{aligned}
&c,\quad min(l^*,t^*,r^*,b^* ) >0\\
&0,\quad otherwise
\end{aligned}
\right.
\end{equation}
\begin{equation}
where \quad c=\sqrt{ \frac{ min(l^*, r^*)}{max(l^*, r^*)} \times \frac{ min(t^*, b^*)}{max(t^*, b^*)}}.
\end{equation}
In this way, the center-ness targets range from 0 to 1. For the positive samples in the score map, the smaller distances between the position of positive samples and the tracking target's center, the greater response that they can get in the center-ness map. We consider the points that close to the center of the target has a better regression effect. To some extent, the center-ness sub-branch can also alleviate the impact of partial occlusion. For the negative samples in the score map, they have no bounding box regression, so we simply set them to 0 in the center-ness target.
\subsection{Hard Negative Samples Emphasis}
\label{HE}
We observe that many Siamese-based trackers are not discriminative for objects belonging to the same category, which leads to tracking failures of Siamese-based algorithms. To address this issue, we analyze the reasons leading to the problem and put forward our improvement.

Firstly, modern deep networks are introduced into the tracking community one after another \cite{zhang2019deeper, li2019siamrpn++}.
However, deep networks are pre-trained on ImageNet \cite{krizhevsky2012imagenet} for image classification. Because of rich semantic information, deep convolutional features are more beneficial for image classification. By contrast, shallow features contain low-level information and their high spatial resolution are more suitable for visual tracking. Siamese-based trackers simply use image pairs to fine-tune the networks. The output features contain most of the semantic information and ignore important detailed information to discriminate intra-class objects.

Secondly, there is no efficient method to constrain hard negative samples in the training phase. We classify samples on the instance \textbf{x} into three classes: positive samples, hard negative samples and easy negative samples. Hard negative samples are composed of cluster background and other similar objects. Existing tracking methods equally treat hard/easy negative samples, which makes them unable to distinguish hard negative samples and the performance is limited.

\begin{figure}[t]
\centering
\includegraphics[scale=0.33]{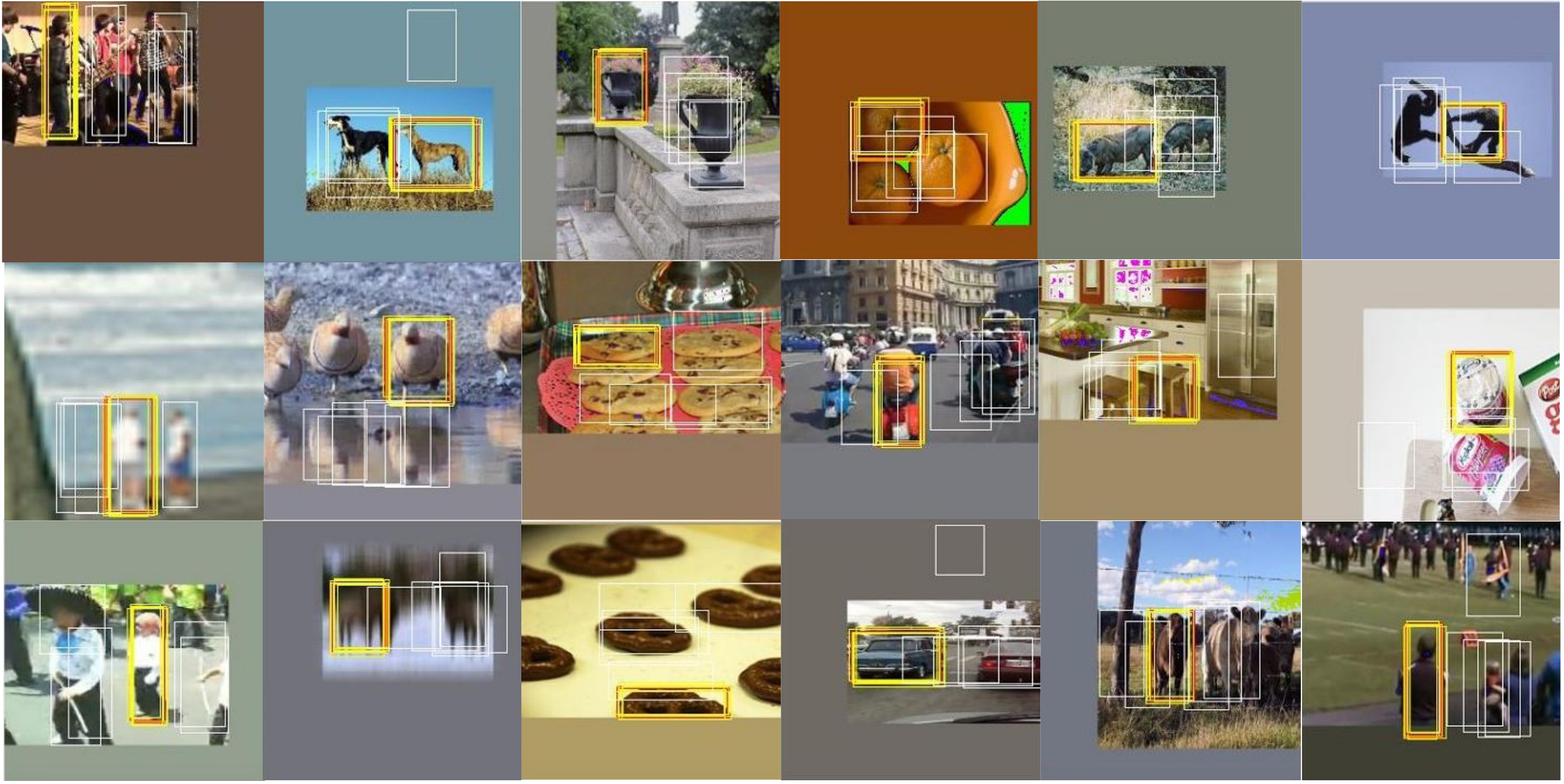}
\caption{Illustration of hard negative samples in training datasets. Bounding boxes in red, yellow and white denote ground-truth, positive samples and hard negative samples respectively.}
\label{4}
\end{figure}

Based on the above analyses and inspired by person re-identification \cite{yang2017person, yang2017enhancing, yang2018person}, we put forward our hard negative samples emphasis method to solve the aforementioned problem. Person re-identification is a task to find the same person across cameras, which commonly uses neural networks to extract embedding vectors that represent the appearance of that object. Then a distance metric is adopted to measure their similarity. For the tracking task, we intend to adopt the same idea to keep the discrimination of positive samples and hard negative samples. As illustrated in Figure \ref{4}, many sequences of training sets contain hard negative samples that should be constrained in the training phase. To be specific, we first select related bounding boxes according to the score ranking in the score map. Non-Maximal Suppression (NMS) \cite{ren2015faster} is employed to eliminate redundant bounding boxes. In this way, we can obtain the candidate bounding boxes that include positive samples and hard negative samples. Specially, positive samples are defined as the candidates which have $IoU > T_{h}$ with the corresponding ground-truth. Hard negative ones are defined as the candidates which satisfy $IoU < T_{l}$. To avoid a bad training effect caused by the imbalance of positive samples and hard negative samples, we adopt a random shift operation of the ground-truth to generate positive samples.
In this way, we can obtain high-quality RoIs in the instance x. In Figure \ref{4}, bounding boxes in yellow represent positive samples and the bounding boxes in white represent hard negative samples.

Considering that deeper features contain less detailed information, our goal is forcing deep features to keep the discrimination of positive samples and hard negative samples. After getting the selected bounding boxes, a RoI Align \cite{he2017mask} layer crops the region on exemplar z or instance x, resulting in the output feature maps of a pre-determined size. For the exemplar branch of Siamese network, a RoI Align \cite{he2017mask} layer takes the fused features  $Cat(\varphi_3(z), \varphi_4(z))$ and the ground-truth bounding box as input, the output is the regional feature with the size of $5 \times 5 \times C$. Then two convolutional layers adjust it to a vector, namely \emph{exemplar vector}. For the instance branch, we use the same way to generate \emph{positive vectors} and \emph{negative vectors}. Here we consider it as a recognition problem. By the constraint of the loss function, we intend to make the distance between \emph{exemplar vector} and \emph{positive vectors} as small as possible, meanwhile, the distances between \emph{exemplar vector} and \emph{negative vectors} are enabled to be as long as possible. A contrastive loss \cite{hadsell2006dimensionality} is adopted here and formulated as follows:
\begin{equation}
L_{contra}= \frac{1}{N} \sum_{i=1}^{N}[y_i \cdot d_i^2 + (1-y_i) \cdot max(0, m-d_i)^2]
\end{equation}
where $d_i$ denotes the Euclidean distance between \emph{exemplar vector} and \emph{positive}/\emph{negative vectors}, $y_i \in \{0,1\}$ represents their corresponding labels. For positive vectors, we set $y_i = 1$, otherwise, $y_i = 0$. Here, $m$ denotes a margin and is equal to 2. $N$ is the number of selected RoIs.
\subsection{Offline Training}
The whole framework is composed of an anchor-free Siamese tracker and a hard negative samples emphasis network. We have already detailed anchor-free Siamese tracker in Section \ref{section3.2}. For the classification branch, we adopt a focal loss \cite{lin2017focal} to restrain the score map, and train the center-ness map with a Binary Cross Entropy (BCE) loss. For the regression branch, the IoU loss \cite{yu2016unitbox} is employed which is formulated as:
\begin{equation}
\label{equation8}
L_{reg} = \frac{1}{N_{p}} \sum_{i=1}^{N_{p}}-ln(\frac{P_i \cap G}{P_i \cup G})
\end{equation}
where $P$ denotes the predicted bounding boxes and $G$ denotes the ground-truth. $Np$ is the number of positive samples in the score map. $\cap$ and $\cup$ represent intersection operator and union operator respectively. We optimize the final loss function as follows,
\begin{equation}
L=\lambda_1 L_{sco}+\lambda_2 L_{cen}+\lambda_3 L_{reg}+\lambda_4 L_{contra}
\end{equation}
where $L_{sco}$ is the focal loss \cite{lin2017focal} in charge of foreground-background classification, $L_{cen}$ denotes a Binary Cross Entropy (BCE) for the center-ness map. $L_{reg}$ represents the IoU loss \cite{yu2016unitbox} as shown in Eq .(\ref{equation8}) and $L_{contra}$ denotes the contrastive loss to enlarge the distances between positive vectors and hard negative vectors. The hyper-parameters are set to $\lambda_1=\lambda_2=\lambda_3=1, \lambda_4=0.1$.
\subsection{Online Tracking}
We first get the exemplar z from the first frame, which is not updated throughout the whole tracking process. Consequently, one branch of Siamese network takes the exemplar z as input. The instance x, which is cropped from the current frame, is taken as the input of another branch. As illustrated in Figure \ref{2}, there are three FCOS heads in the framework, we can obtain a score map (s), a center-ness map (c) and a regression (r) by each FCOS head. The output can be obtained by a weighted summation,
\begin{equation}
S=\sum_{i=2}^{4} w_i^s \cdot s_i,\quad C=\sum_{i=2}^{4} w_i^c \cdot c_i, \quad R=\sum_{i=2}^{4} w_i^r \cdot r_i.
\end{equation}
where $w_i^s$, $w_i^c$, $w_i^r$, are trainable parameters. Then, we exert a softmax function in the score map (S) for foreground-background discrimination and perform an element-wise multiplication between the score map and the center-ness map to get the final score map. To penalize large displacements and suppress the large change in size and aspect ratio, we also exert a cosine window and a scale change penalty \cite{li2018high} to the final score map. The scale change penalty is formulated as:
\begin{equation}
 penalty = exp(k \cdot max(\frac{r}{r^\prime},\frac{r^\prime}{r}) \cdot max(\frac{s}{s^\prime},\frac{s^\prime}{s})).
\end{equation}
where $k$ is a hyper-parameter, $r$ denotes the aspect ratio of the predicted bounding boxes in the current frame and $r^\prime$ is that in the previous frame, $s$ is computed by
\begin{equation}
s = \sqrt{(w + p) \cdot (h + p})
\end{equation}
and $s^\prime$ is that in the previous frame, here $w$ and $h$ denote width and height of the target. $p$ is equal to $(w+h)/2$.
After that, the coordinate of the highest score value in the final score map is exactly the position of the tracking target and is the position of the selected bounding box.
\section{Experiments}
In this section, we first introduce the implementation details of our method. Then, we compare our approach with state-of-the-art trackers on six commonly used benchmark datasets, including GOT-10k \cite{huang2019got}, UAV123 \cite{mueller2016benchmark}, VOT2016 \cite{Kristan2016a}, VOT2018 \cite{Kristan2018a}, VOT2019 \cite{Kristan2019a}, LaSOT \cite{fan2019lasot}. Finally, we perform ablation studies to demonstrate the effectiveness of our method.
\subsection{Implementation Details}
\textbf{Data Preprocessing.} \quad We train the network on six large training datasets including YouTube-BB \cite{real2017youtube}, COCO \cite{lin2014microsoft}, ImageNet-VID \cite{krizhevsky2012imagenet}, ImageNet-DET \cite{krizhevsky2012imagenet}, GOT-10k \cite{huang2019got}, LaSOT \cite{fan2019lasot}. The size of exemplar z is $127 \times 127 \times 3$ and the size of instance x is $255 \times 255 \times 3$. In order to eliminate the impact of the convolutional layers' padding operation and simulate the real tracking scenarios, a random translation ($\pm 64$ pixel) is exerted to the target in the instance x. During training, we select pairs of cropped images that belong to the same sequences, and the interval of image pairs is less than 100 frames for ImageNet-VID \cite{krizhevsky2012imagenet}, LaSOT \cite{fan2019lasot}, GOT-10k \cite{huang2019got} and less than 3 frames for Youtube-BB \cite{real2017youtube}. Some data augmentation methods like scaling and color change are also employed for better training effect.
\\
\textbf{Training Details.} \quad We use a modified ResNet-50 \cite{he2016deep} pre-trained on ImageNet \cite{krizhevsky2012imagenet} as our backbone. In the training stage,  We train the model for 20 epochs with mini-batches of size 28. For the first 10 epochs, we freeze the parameters of backbone and only fine-tune the parameters of three FCOS \cite{tian2019fcos} heads. For the last 10 epochs, the parameters of \emph{block2}, \emph{block3} and \emph{block4} are unfrozen. The base learning rate is $5 \times 10^{-3}$ and we use a warm-up learning rate of $10^{-3}$ for the first 5 epochs. For the last 15 epochs,  the learning rate is decreased exponentially at each epoch from $5 \times 10^{-3}$ to $5 \times 10^{-4}$. We train the network by stochastic gradient descent (SGD) with a weight decay of $10^{-4}$ and a momentum of $0.9$. For hard negative samples emphasis method, the threshold of classifying positive samples and hard negative samples are set to $T_{h} = 0.8$ and $T_{l}=0.3$. The threshold of NMS is equal to $0.7$. Our training experiments are implemented using PyTorch on PC with an Intel i9-7900X and four NVIDIA Titan XP GPUs.\\
\begin{figure}[ht]
\centering
\includegraphics[scale=0.35]{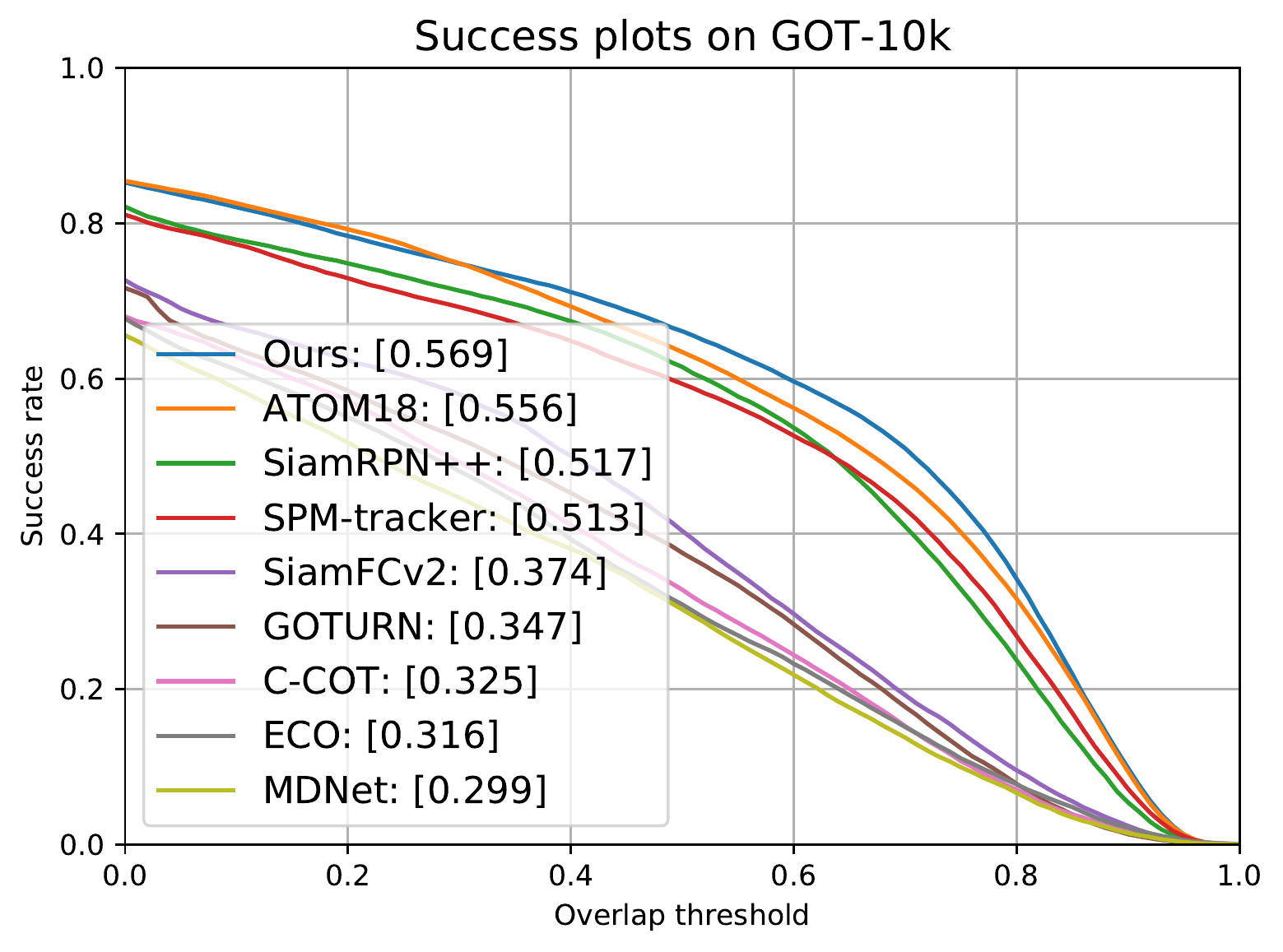}
\caption{ Comparison of state-of-the-art trackers on GOT-10k \cite{huang2019got} in terms of success rate.}
\label{5}
\end{figure}
\begin{table*}[t]
\small
\centering
\caption{Performance comparisons on GOT-10k \cite{huang2019got} benchmark. AO: average overlap; SR: success rate wiht different thresholds of 0.5 and 0.75. The best two results are highlighted in \textcolor[rgb]{1.00,0.00,0.00}{red} and \textcolor[rgb]{0.00,0.07,1.00}{blue} respectively.}
\setlength{\tabcolsep}{3mm}{
\begin{tabular}{c| c c c c c c c c c}
\hline
\hline
Trackers                  & MDNet & ECO  & CCOT & GOTURN & SiamFCv2 &SPM-traker&SiamRPN++ & ATOM                            &Ours\\
                          & \cite{nam2016learning} &\cite{danelljan2017eco} &\cite{danelljan2016beyond} &\cite{held2016learning}&\cite{bertinetto2016fully} &\cite{wang2019spm} &\cite{li2019siamrpn++} &\cite{danelljan2019atom} & \\\hline
AO $\uparrow$             &0.299  &0.316 &0.325 &0.347   &0.348   &0.513     &0.517     &\textcolor[rgb]{0,0.07,1}{0.556} &\textcolor[rgb]{1,0,0}{0.569} \\
SR$_{0.5}$ $\uparrow$     &0.303  &0.309 &0.328 &0.375   &0.353   &0.593     &0.615     &\textcolor[rgb]{0,0.07,1}{0.634} &\textcolor[rgb]{1,0,0}{0.661} \\
SR$_{0.75} $ $\uparrow$   &0.099  &0.111 &0.107 &0.124   &0.098   &0.359     &0.329     &\textcolor[rgb]{0,0.07,1}{0.402} &\textcolor[rgb]{1,0,0}{0.439}\\
\hline
\hline
\end{tabular}}
\label{got10ktable}
\end{table*}
\subsection{Results on GOT-10k}
GOT-10k \cite{huang2019got}, which consists of 10,000 short sequences, is a large tracking database that offsets a wild converge of common moving objects in the wild. Huang et al. \cite{huang2019got} select 180 videos to form the test subset for evaluation. They take average overlap (AO) and success rate (SR) as the performance indicators. AO denotes the average of overlaps between all ground-truth and estimated bounding boxes, while SR measures the percentage of successful tracked frames where the overlaps exceed a threshold (SR$_{0.5}$ and SR$_{0.75}$). For evaluation, all results are evaluated by the server provided by Huang et al. \cite{huang2019got}.

We can obtain outstanding performance on GOT-10k. Success plot is shown in Figure \ref{5}, and the detailed performance scores of trackers are illustrated in table \ref{got10ktable}. Our method can achieve AO score of 0.569, outperforming the second-best tracker ATOM \cite{danelljan2019atom} with a relative gain of $2.3\%$ and exceeding the baseline method (SiamRPN++ \cite{li2019siamrpn++}) by $10\%$ improvement . As for the success rate, our tracker can get SR$_{0.5}$ and SR$_{0.75}$ scores of 0.661 and 0.439 while tracking with a real-time speed.
\begin{table}[t]
\centering
\small
\caption{Comparison of state-of-the-art trackers on VOT2016 \cite{Kristan2016a} and VOT2018 \cite{Kristan2018a} in terms of Accuracy, Robustness, EAO and speed. The best two results are highlighted in \textcolor[rgb]{1.00,0.00,0.00}{red} and \textcolor[rgb]{0.00,0.07,1.00}{blue} respectively.}
\setlength{\tabcolsep}{0.9mm}{
\begin{tabular}{c|ccc|ccc|c}
\hline
\hline
\multicolumn{1}{c}{ \multirow{2}*{Trackers} }& \multicolumn{3}{|c|}{VOT2016} & \multicolumn{3}{c|}{VOT2018} & \multicolumn{1}{c}{ \multirow{2}*{FPS} }\\
\cline{2-7}
\multicolumn{1}{c|}{}&A$\uparrow$&R$\downarrow$&EAO$\uparrow$&A$\uparrow$&R$\downarrow$&EAO$\uparrow$\\
\hline
\hline
ECO\cite{danelljan2017eco}             &0.54 &0.20&0.374    &0.484&0.276&0.280&8\\
SiamVGG\cite{li2019siamvgg}         &0.564& -  &0.351    &0.531&0.318&0.286&33\\
SA\_Siam\_R\cite{he2018twofold}     & 0.54& -  &0.291    &0.566&0.258&0.337& 50  \\
TADT\cite{li2019target}            &0.55 & -  &0.299    &-&-&-     &33.7\\
C-RPN\cite{fan2019siamese}           &0.594& -  &0.363    & -  & - & 0.289 & 36\\
SiamDW-rpn\cite{zhang2019deeper}          &   - &   -&0.376    &- &-&0.294 & -  \\
SiamRPN\cite{li2018high}         &0.56 &0.26&0.344    & 0.49   &  0.46  &  0.244 &\textcolor[rgb]{1.00,0.00,0.00}{200}    \\
SPM-Tracker\cite{wang2019spm}     &0.62 &0.21&0.434    & 0.58   &  0.30  &  0.338  & 120 \\
UPDT\cite{bhat2018unveiling}            &- &-&-   &0.536&0.184&0.378   &86\\
DaSiamRPN\cite{zhu2018distractor}       &0.61 &0.22&0.411    & 0.586   &  0.276  &  0.383 &\textcolor[rgb]{0.00,0.07,1.00}{160}    \\
LADCF\cite{xu2019learning}           & -   & -  & -       & 0.503   &  \textcolor[rgb]{1.00,0.00,0.00}{0.159}  &  0.389 & -     \\
ATOM\cite{danelljan2019atom}            & -   &  - & -       & 0.590  &  0.204 &  0.401  & 30 \\
SiamRPN++\cite{li2019siamrpn++}       & \textcolor[rgb]{1.00,0.00,0.00}{0.637}   & \textcolor[rgb]{0.00,0.07,1.00}{0.177} & \textcolor[rgb]{0.00,0.07,1.00}{0.478}       & \textcolor[rgb]{1.00,0.00,0.00}{0.600}  &  0.234 &  \textcolor[rgb]{0.00,0.07,1.00}{0.414}  & 35 \\
Ours            &\textcolor[rgb]{0.00,0.07,1.00}{0.635} &\textcolor[rgb]{1.00,0.00,0.00}{0.158} &\textcolor[rgb]{1.00,0.00,0.00}{0.487}    & \textcolor[rgb]{0.00,0.07,1.00}{0.596}  & \textcolor[rgb]{0.00,0.07,1.00}{0.183} &  \textcolor[rgb]{1.00,0.00,0.00}{0.433}  &37    \\
\hline
\hline
\end{tabular}}
\label{vot2016and2018}
\end{table}
\begin{table}[t]
\small
\centering
\caption{Comparison of state-of-the-art trackers on VOT2019 \cite{Kristan2019a} in terms of EAO, Accuracy and Robustness.}
\setlength{\tabcolsep}{1.2mm}{
\begin{tabular}{c| c c c c c c}
\hline
\hline
Trackers         & SA\_Siam\_R &SPM       &SiamRPN++ &SiamMask   &Ours\\ \hline
EAO $\uparrow$   &   0.253   &0.275       &0.285     &\textcolor[rgb]{0.00,0.07,1.00}{0.287}      &\textcolor[rgb]{1,0,0}{0.303} \\
A   $\uparrow$   &   0.559   &0.577       &\textcolor[rgb]{0.00,0.07,1.00}{0.599}     &0.594        &\textcolor[rgb]{1.00,0.00,0.00}{0.600} \\
R   $\downarrow$ &   0.492   &0.507       &0.482     &\textcolor[rgb]{0.00,0.07,1.00}{0.461}        &\textcolor[rgb]{1.00,0.00,0.00}{0.391} \\
\hline
\hline
\end{tabular}}
\label{vot2019table}
\end{table}

\subsection{Results on VOT}
Visual Object Tracking (VOT) dataset \cite{Kristan2018a} consists of 60 challenging sequences and some sequences are updated annually. We usually evaluate the performance of trackers with the official VOT toolkit.
The performance is evaluated in terms of accuracy (A) and robustness (R), which respectively denote the average overlap over successfully tracked frames and failure rate. Expected Average Overlap (EAO), which merges both accuracy and robustness, is always used to ranking test.

In table \ref{vot2016and2018} and table \ref{vot2019table}, we report the experiment results on VOT2016 \cite{Kristan2016a}, VOT2018 \cite{Kristan2018a} and VOT2019 \cite{Kristan2019a}. Compared with state-of-the-art trackers, our method can get outstanding performance. For a fair comparison, our test environment is the same as SiamRPN++ \cite{li2019siamrpn++}.
On VOT2016, we can get EAO score of 0.487 that outperforms the baseline method SiamRPN++ \cite{li2019siamrpn++} with a relative gain of $1.8\%$ in terms of EAO. On VOT2018,  the EAO score of our tracker is higher than SiamRPN++ \cite{li2019siamrpn++} by $4.6\%$.
Furthermore, on VOT2019, our method can surpass SiamRPN++ \cite{li2019siamrpn++} in each indicators.
\begin{figure}[t]
\centering
\includegraphics[scale=0.45]{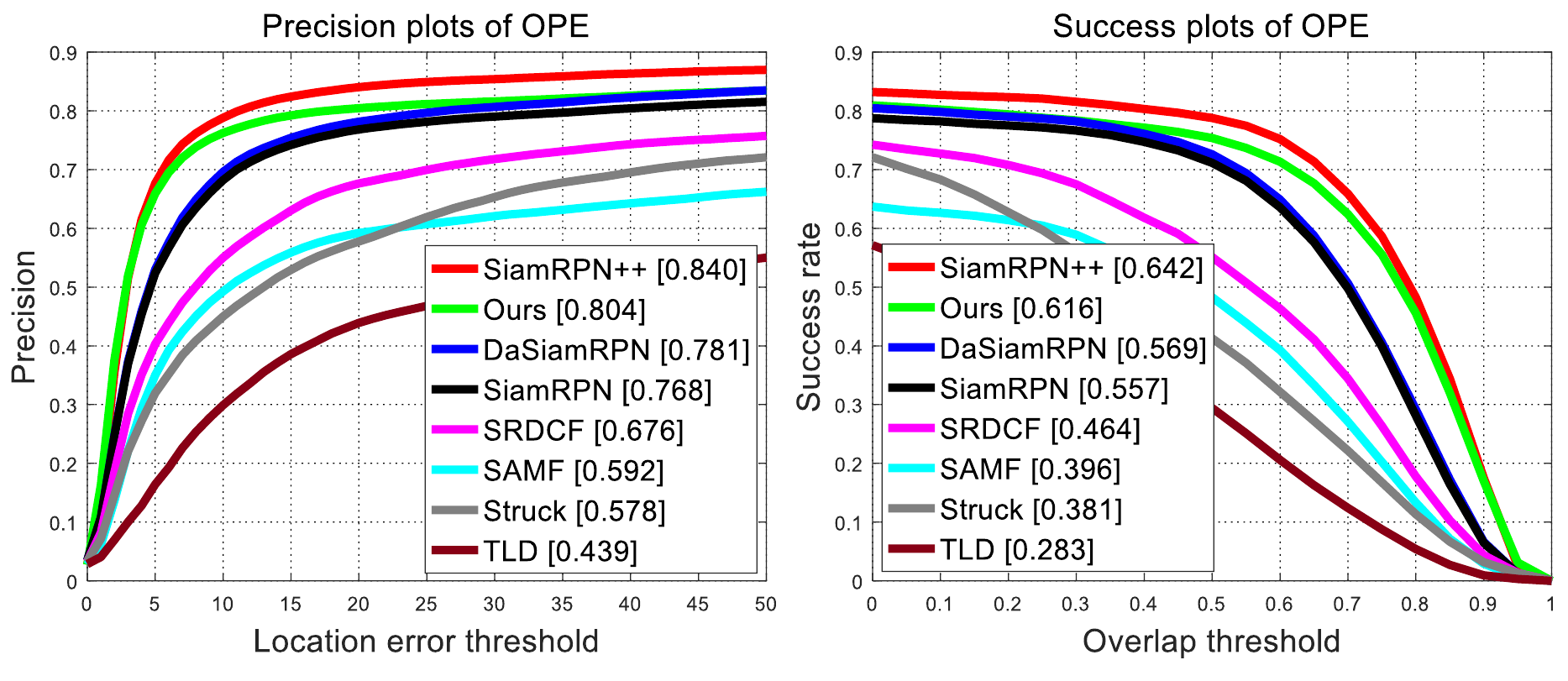}
\caption{ Comparisons with state-of-the-art tracking approaches on UAV123 \cite{mueller2016benchmark} in terms of precision and success rate.}
\label{uav123plot}
\end{figure}

\subsection{Results on UAV123}
UAV123 \cite{mueller2016benchmark} consists of 123 aerial video sequences comprising more than 110K frames, which takes precision and success rate as indicators for evaluation. The comparison of the proposed method with representative trackers (SiamRPN++ \cite{li2019siamrpn++}, DaSiamRPN \cite{zhu2018distractor}, SiamRPN \cite{li2018high}, SRDCF \cite{danelljan2015learning}, SAMF \cite{li2014scale}, Struck \cite{hare2015struck}, TLD \cite{Kalal2011Tracking}) is illustrated in Figure \ref{uav123plot}. We can see that our approach performs the two-best in this dataset in terms of precision and success rate. Our tracker achieves a precision score of 0.805 and a success score of 0.615, which are close to the best results.
\begin{figure}[t]
\centering
\includegraphics[scale=0.44]{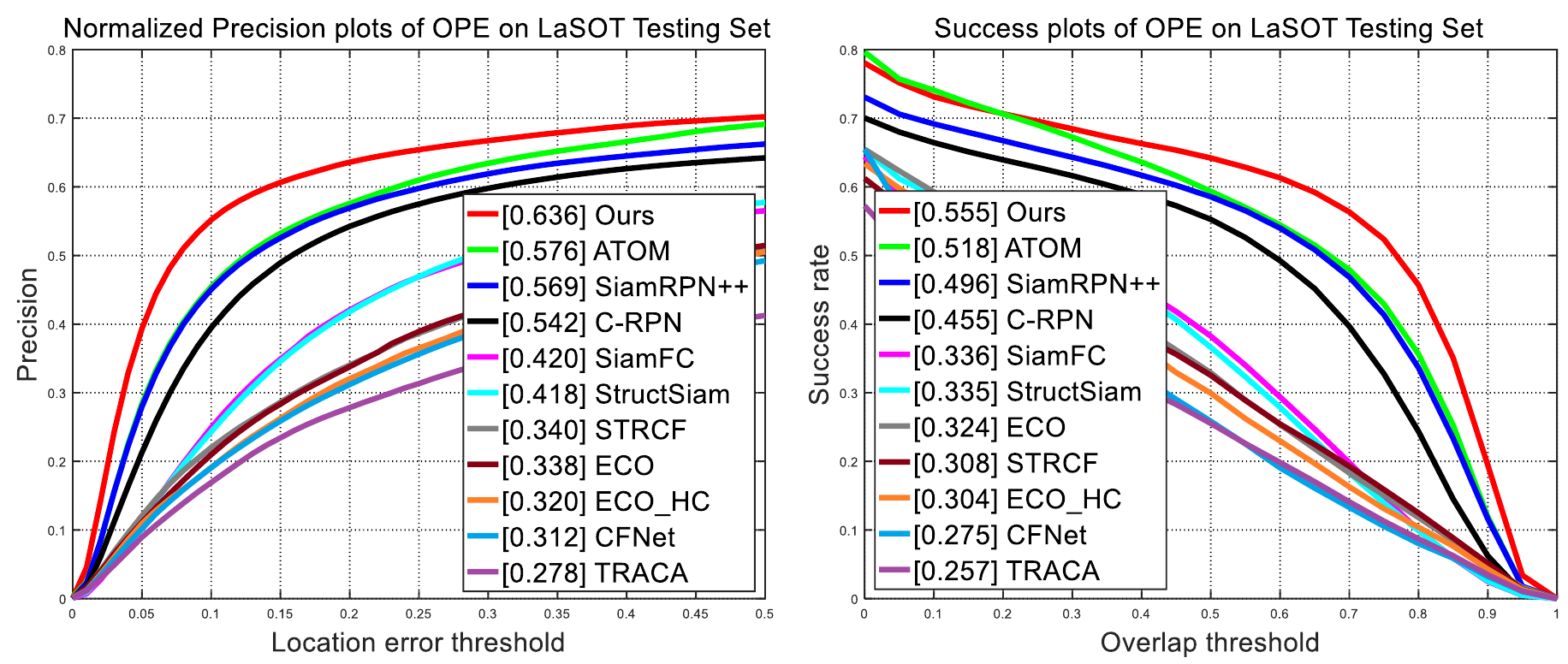}
\caption{ Comparison of state-of-the-art trackers on LaSOT \cite{fan2019lasot} in terms of normalized precision and success rate.}
\label{lasotplot}
\end{figure}
\subsection{Results on LaSOT}
The Large-scale Single Object Tracking dataset (LaSOT) \cite{fan2019lasot} is composed of  1,400 sequences and more than 3.5M frames that are labeled with high-quality dense annotations including 70 categories. The test subset of LaSOT contain 280 long-term videos. The average frame length of this dataset is more than 2,500 frames, so the evaluation results are convincing and truly reflect the performance of various trackers. LaSOT takes normalized precision and success rate as indicators for evaluation. Our algorithm is compared with other 14 trackers (SiamRPN++ \cite{li2019siamrpn++}, ATOM \cite{danelljan2019atom}, C-RPN \cite{fan2019siamese}, SiamFC \cite{bertinetto2016fully}, StructSiam \cite{zhang2018structured}, ECO \cite{danelljan2017eco}, CFNet \cite{Valmadre2017End}, TRACA \cite{Choi2018Context}, STRCF \cite{li2018learning}). The normalized precision plot and success plot are illustrated in Figure \ref{lasotplot}.
Our tracker performs significantly better than the baseline model (SiamRPN++ \cite{li2019siamrpn++}) by almost $11.7\%$ in normalized precision and $11.8\%$ in success.
\subsection{Ablation Experiment}
\begin{figure}[t]
\centering
\includegraphics[scale=0.65]{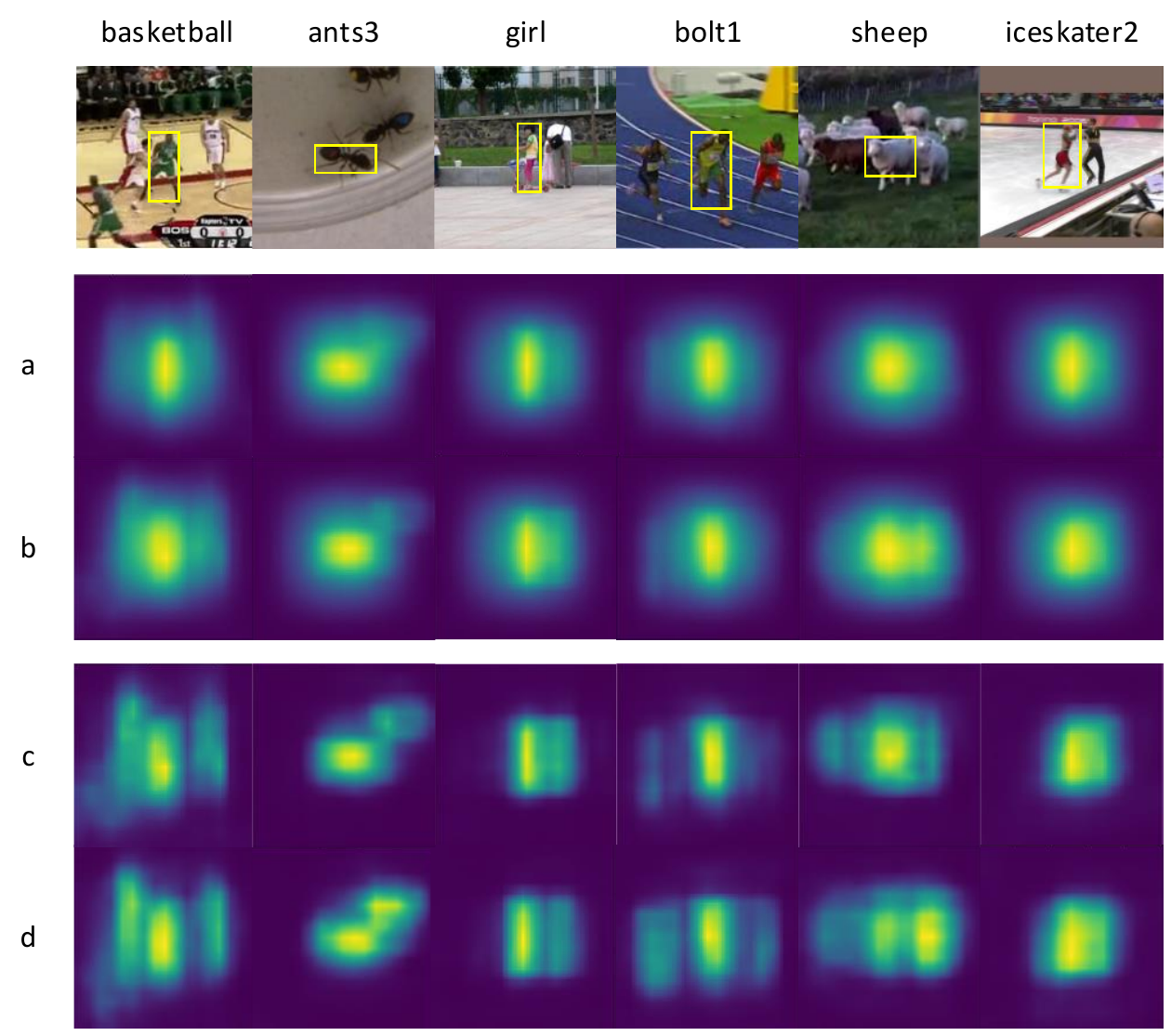}
\caption{Visualization of score maps. The first row shows the original images of sequences. And (a) and (b) denote final score maps with penalty, (c) and (d) show score maps without penaty, (a) and (c) represent our model with hard negative samples emphasis. Bounding boxes in yellow denote tracking targets.}
\label{heatmap}
\end{figure}

\textbf{Analysis of hard negative samples emphasis:} We evaluate our approach in different situations. Table \ref{ablationgot10k} shows the results using GOT-10k \cite{huang2019got} dataset under the measure of three key performance indicators. In the first column of table \ref{ablationgot10k}, `\emph{Baseline}' represents our baseline method SiamRPN++ \cite{li2019siamrpn++}.  `\emph{Ours w/o HE}' denotes the original model without hard negative samples emphasis and `\emph{Ours}' represents the complete model that we put forward. Table \ref{ablationgot10k} demonstrates `\emph{Ours w/o HE}' outperforms `\emph{Baseline}' with a relative gain of $5.4\%$ in terms of AO. By adding hard negative samples emphasis method, we improve the AO score on GOT-10K from 0.545 to 0.569. Furthermore, we select some representative sequences with hard negative samples from VOT2018 \cite{Kristan2018a} and visualise their score maps. As shown in Figure \ref{heatmap}. Obviously, the score map of `\emph{Our}' shows better discriminability than that of `\emph{Ours w/o HE}'.
\begin{table}[ht]
  \caption{ablation experiment on GOT-10k \cite{huang2019got}}
  \setlength{\tabcolsep}{3.5mm}{
  \begin{tabular}{c | c c c}
    \hline
    \hline
    &AO$\uparrow$& SR$_{0.5}$$\uparrow$  & SR$_{0.75}$ $\uparrow$ \\
    \hline
    Baseline& 0.517  & 0.615 & 0.329\\
    Ours w/o HE& 0.545 & 0.638 & 0.388\\
    Ours& 0.569 & 0.661 & 0.439\\
    \hline
    \hline
\end{tabular}}
\label{ablationgot10k}
\end{table}
\begin{table}[ht]
  \caption{Influence of the number of RoIs}
  \setlength{\tabcolsep}{3.5mm}{
  \begin{tabular}{c | c c c}
    \hline
    \hline
    RoIs' number & AO$\uparrow$& SR$_{0.5}$$\uparrow$  & SR$_{0.75}$ $\uparrow$ \\
    \hline
    5& 0.548 & 0.642 & 0.401\\
     11& 0.569 & 0.661 & 0.439\\
    21 & 0.556 & 0.648 & 0.432\\
    31& 0.538  & 0.631 & 0.385\\
    \hline
    \hline
\end{tabular}}
\label{ablationvot}
\end{table}
\\\textbf{Impact of the selected RoIs' number:} We have performed an ablation study on different numbers of RoIs. These selected RoIs include k bounding boxes of positive samples, k bounding boxes of hard negative samples and one ground-truth bounding box. As shown in Table \ref{ablationvot}, we can get the best result when the number of RoIs is equal to 11. If there are too many selected RoIs, the performance of trackers will decrease.
\section{Conclusions}
In this paper, we perform a novel bounding box prediction framework with a per-pixel prediction fashion to train Siamese network for visual tracking. Compared with trackers based on RPN, our anchor-free approach can generate a more precise bounding box and reduce the number of hyper-parameters. Furthermore, we have a deep analysis of the tracking failure caused by hard negative samples and propose a hard negative samples emphasis method to improve the discriminative power of Siamese networks. The evaluation results on frequently-used tracking benchmarks demonstrate that our proposed tracker can improve the performance and achieve real-time tracking.
\begin{acks}
This work was supported by the National Science Fund of China under Grants (61771079) and Chongqing Youth Talent Program.
\end{acks}
\bibliographystyle{ACM-Reference-Format}
\bibliography{sample-base}


\begin{thebibliography}{51}


\ifx \showCODEN    \undefined \def \showCODEN     #1{\unskip}     \fi
\ifx \showDOI      \undefined \def \showDOI       #1{#1}\fi
\ifx \showISBNx    \undefined \def \showISBNx     #1{\unskip}     \fi
\ifx \showISBNxiii \undefined \def \showISBNxiii  #1{\unskip}     \fi
\ifx \showISSN     \undefined \def \showISSN      #1{\unskip}     \fi
\ifx \showLCCN     \undefined \def \showLCCN      #1{\unskip}     \fi
\ifx \shownote     \undefined \def \shownote      #1{#1}          \fi
\ifx \showarticletitle \undefined \def \showarticletitle #1{#1}   \fi
\ifx \showURL      \undefined \def \showURL       {\relax}        \fi
\providecommand\bibfield[2]{#2}
\providecommand\bibinfo[2]{#2}
\providecommand\natexlab[1]{#1}
\providecommand\showeprint[2][]{arXiv:#2}

\bibitem[\protect\citeauthoryear{Bertinetto, Valmadre, Henriques, Vedaldi, and
  Torr}{Bertinetto et~al\mbox{.}}{2016}]%
        {bertinetto2016fully}
\bibfield{author}{\bibinfo{person}{Luca Bertinetto}, \bibinfo{person}{Jack
  Valmadre}, \bibinfo{person}{Joao~F Henriques}, \bibinfo{person}{Andrea
  Vedaldi}, {and} \bibinfo{person}{Philip~HS Torr}.}
  \bibinfo{year}{2016}\natexlab{}.
\newblock \showarticletitle{Fully-convolutional siamese networks for object
  tracking}. In \bibinfo{booktitle}{\emph{European conference on computer
  vision}}. Springer, \bibinfo{pages}{850--865}.
\newblock


\bibitem[\protect\citeauthoryear{Bhat, Johnander, Danelljan, Shahbaz~Khan, and
  Felsberg}{Bhat et~al\mbox{.}}{2018}]%
        {bhat2018unveiling}
\bibfield{author}{\bibinfo{person}{Goutam Bhat}, \bibinfo{person}{Joakim
  Johnander}, \bibinfo{person}{Martin Danelljan}, \bibinfo{person}{Fahad
  Shahbaz~Khan}, {and} \bibinfo{person}{Michael Felsberg}.}
  \bibinfo{year}{2018}\natexlab{}.
\newblock \showarticletitle{Unveiling the power of deep tracking}. In
  \bibinfo{booktitle}{\emph{Proceedings of the European Conference on Computer
  Vision (ECCV)}}. \bibinfo{pages}{483--498}.
\newblock


\bibitem[\protect\citeauthoryear{Choi, Chang, Fischer, Yun, and Jin}{Choi
  et~al\mbox{.}}{2018}]%
        {Choi2018Context}
\bibfield{author}{\bibinfo{person}{Jongwon Choi}, \bibinfo{person}{Hyung~Jin
  Chang}, \bibinfo{person}{Tobias Fischer}, \bibinfo{person}{Sangdoo Yun},
  {and} \bibinfo{person}{Young~Choi Jin}.} \bibinfo{year}{2018}\natexlab{}.
\newblock \showarticletitle{Context-Aware Deep Feature Compression for
  High-Speed Visual Tracking}. In \bibinfo{booktitle}{\emph{IEEE Conference on
  Computer Vision and Pattern Recognition}}.
\newblock


\bibitem[\protect\citeauthoryear{Danelljan, Bhat, Khan, and Felsberg}{Danelljan
  et~al\mbox{.}}{2019}]%
        {danelljan2019atom}
\bibfield{author}{\bibinfo{person}{Martin Danelljan}, \bibinfo{person}{Goutam
  Bhat}, \bibinfo{person}{Fahad~Shahbaz Khan}, {and} \bibinfo{person}{Michael
  Felsberg}.} \bibinfo{year}{2019}\natexlab{}.
\newblock \showarticletitle{Atom: Accurate tracking by overlap maximization}.
  In \bibinfo{booktitle}{\emph{Proceedings of the IEEE Conference on Computer
  Vision and Pattern Recognition}}. \bibinfo{pages}{4660--4669}.
\newblock


\bibitem[\protect\citeauthoryear{Danelljan, Bhat, Shahbaz~Khan, and
  Felsberg}{Danelljan et~al\mbox{.}}{2017}]%
        {danelljan2017eco}
\bibfield{author}{\bibinfo{person}{Martin Danelljan}, \bibinfo{person}{Goutam
  Bhat}, \bibinfo{person}{Fahad Shahbaz~Khan}, {and} \bibinfo{person}{Michael
  Felsberg}.} \bibinfo{year}{2017}\natexlab{}.
\newblock \showarticletitle{Eco: Efficient convolution operators for tracking}.
  In \bibinfo{booktitle}{\emph{Proceedings of the IEEE conference on computer
  vision and pattern recognition}}. \bibinfo{pages}{6638--6646}.
\newblock


\bibitem[\protect\citeauthoryear{Danelljan, Hager, Shahbaz~Khan, and
  Felsberg}{Danelljan et~al\mbox{.}}{2015}]%
        {danelljan2015learning}
\bibfield{author}{\bibinfo{person}{Martin Danelljan}, \bibinfo{person}{Gustav
  Hager}, \bibinfo{person}{Fahad Shahbaz~Khan}, {and} \bibinfo{person}{Michael
  Felsberg}.} \bibinfo{year}{2015}\natexlab{}.
\newblock \showarticletitle{Learning spatially regularized correlation filters
  for visual tracking}. In \bibinfo{booktitle}{\emph{Proceedings of the IEEE
  international conference on computer vision}}. \bibinfo{pages}{4310--4318}.
\newblock


\bibitem[\protect\citeauthoryear{Danelljan, Robinson, Khan, and
  Felsberg}{Danelljan et~al\mbox{.}}{2016}]%
        {danelljan2016beyond}
\bibfield{author}{\bibinfo{person}{Martin Danelljan}, \bibinfo{person}{Andreas
  Robinson}, \bibinfo{person}{Fahad~Shahbaz Khan}, {and}
  \bibinfo{person}{Michael Felsberg}.} \bibinfo{year}{2016}\natexlab{}.
\newblock \showarticletitle{Beyond correlation filters: Learning continuous
  convolution operators for visual tracking}. In
  \bibinfo{booktitle}{\emph{European conference on computer vision}}. Springer,
  \bibinfo{pages}{472--488}.
\newblock


\bibitem[\protect\citeauthoryear{Fan, Lin, Yang, Chu, Deng, Yu, Bai, Xu, Liao,
  and Ling}{Fan et~al\mbox{.}}{2019}]%
        {fan2019lasot}
\bibfield{author}{\bibinfo{person}{Heng Fan}, \bibinfo{person}{Liting Lin},
  \bibinfo{person}{Fan Yang}, \bibinfo{person}{Peng Chu}, \bibinfo{person}{Ge
  Deng}, \bibinfo{person}{Sijia Yu}, \bibinfo{person}{Hexin Bai},
  \bibinfo{person}{Yong Xu}, \bibinfo{person}{Chunyuan Liao}, {and}
  \bibinfo{person}{Haibin Ling}.} \bibinfo{year}{2019}\natexlab{}.
\newblock \showarticletitle{Lasot: A high-quality benchmark for large-scale
  single object tracking}. In \bibinfo{booktitle}{\emph{Proceedings of the IEEE
  Conference on Computer Vision and Pattern Recognition}}.
  \bibinfo{pages}{5374--5383}.
\newblock


\bibitem[\protect\citeauthoryear{Fan and Ling}{Fan and Ling}{2019}]%
        {fan2019siamese}
\bibfield{author}{\bibinfo{person}{Heng Fan} {and} \bibinfo{person}{Haibin
  Ling}.} \bibinfo{year}{2019}\natexlab{}.
\newblock \showarticletitle{Siamese cascaded region proposal networks for
  real-time visual tracking}. In \bibinfo{booktitle}{\emph{Proceedings of the
  IEEE Conference on Computer Vision and Pattern Recognition}}.
  \bibinfo{pages}{7952--7961}.
\newblock


\bibitem[\protect\citeauthoryear{Hadsell, Chopra, and LeCun}{Hadsell
  et~al\mbox{.}}{2006}]%
        {hadsell2006dimensionality}
\bibfield{author}{\bibinfo{person}{Raia Hadsell}, \bibinfo{person}{Sumit
  Chopra}, {and} \bibinfo{person}{Yann LeCun}.}
  \bibinfo{year}{2006}\natexlab{}.
\newblock \showarticletitle{Dimensionality reduction by learning an invariant
  mapping}. In \bibinfo{booktitle}{\emph{2006 IEEE Computer Society Conference
  on Computer Vision and Pattern Recognition (CVPR'06)}},
  Vol.~\bibinfo{volume}{2}. IEEE, \bibinfo{pages}{1735--1742}.
\newblock


\bibitem[\protect\citeauthoryear{Hare, Golodetz, Saffari, Vineet, Cheng, Hicks,
  and Torr}{Hare et~al\mbox{.}}{2015}]%
        {hare2015struck}
\bibfield{author}{\bibinfo{person}{Sam Hare}, \bibinfo{person}{Stuart
  Golodetz}, \bibinfo{person}{Amir Saffari}, \bibinfo{person}{Vibhav Vineet},
  \bibinfo{person}{Ming-Ming Cheng}, \bibinfo{person}{Stephen~L Hicks}, {and}
  \bibinfo{person}{Philip~HS Torr}.} \bibinfo{year}{2015}\natexlab{}.
\newblock \showarticletitle{Struck: Structured output tracking with kernels}.
\newblock \bibinfo{journal}{\emph{IEEE transactions on pattern analysis and
  machine intelligence}} \bibinfo{volume}{38}, \bibinfo{number}{10}
  (\bibinfo{year}{2015}), \bibinfo{pages}{2096--2109}.
\newblock


\bibitem[\protect\citeauthoryear{He, Luo, Tian, and Zeng}{He
  et~al\mbox{.}}{2018}]%
        {he2018twofold}
\bibfield{author}{\bibinfo{person}{Anfeng He}, \bibinfo{person}{Chong Luo},
  \bibinfo{person}{Xinmei Tian}, {and} \bibinfo{person}{Wenjun Zeng}.}
  \bibinfo{year}{2018}\natexlab{}.
\newblock \showarticletitle{A twofold siamese network for real-time object
  tracking}. In \bibinfo{booktitle}{\emph{Proceedings of the IEEE Conference on
  Computer Vision and Pattern Recognition}}. \bibinfo{pages}{4834--4843}.
\newblock


\bibitem[\protect\citeauthoryear{He, Gkioxari, Doll{\'a}r, and Girshick}{He
  et~al\mbox{.}}{2017}]%
        {he2017mask}
\bibfield{author}{\bibinfo{person}{Kaiming He}, \bibinfo{person}{Georgia
  Gkioxari}, \bibinfo{person}{Piotr Doll{\'a}r}, {and} \bibinfo{person}{Ross
  Girshick}.} \bibinfo{year}{2017}\natexlab{}.
\newblock \showarticletitle{Mask r-cnn}. In
  \bibinfo{booktitle}{\emph{Proceedings of the IEEE international conference on
  computer vision}}. \bibinfo{pages}{2961--2969}.
\newblock


\bibitem[\protect\citeauthoryear{He, Zhang, Ren, and Sun}{He
  et~al\mbox{.}}{2016}]%
        {he2016deep}
\bibfield{author}{\bibinfo{person}{Kaiming He}, \bibinfo{person}{Xiangyu
  Zhang}, \bibinfo{person}{Shaoqing Ren}, {and} \bibinfo{person}{Jian Sun}.}
  \bibinfo{year}{2016}\natexlab{}.
\newblock \showarticletitle{Deep residual learning for image recognition}. In
  \bibinfo{booktitle}{\emph{Proceedings of the IEEE conference on computer
  vision and pattern recognition}}. \bibinfo{pages}{770--778}.
\newblock


\bibitem[\protect\citeauthoryear{Held, Thrun, and Savarese}{Held
  et~al\mbox{.}}{2016}]%
        {held2016learning}
\bibfield{author}{\bibinfo{person}{David Held}, \bibinfo{person}{Sebastian
  Thrun}, {and} \bibinfo{person}{Silvio Savarese}.}
  \bibinfo{year}{2016}\natexlab{}.
\newblock \showarticletitle{Learning to track at 100 fps with deep regression
  networks}. In \bibinfo{booktitle}{\emph{European Conference on Computer
  Vision}}. Springer, \bibinfo{pages}{749--765}.
\newblock


\bibitem[\protect\citeauthoryear{Huang, Zhao, and Huang}{Huang
  et~al\mbox{.}}{2019}]%
        {huang2019got}
\bibfield{author}{\bibinfo{person}{Lianghua Huang}, \bibinfo{person}{Xin Zhao},
  {and} \bibinfo{person}{Kaiqi Huang}.} \bibinfo{year}{2019}\natexlab{}.
\newblock \showarticletitle{Got-10k: A large high-diversity benchmark for
  generic object tracking in the wild}.
\newblock \bibinfo{journal}{\emph{IEEE Transactions on Pattern Analysis and
  Machine Intelligence}} (\bibinfo{year}{2019}).
\newblock


\bibitem[\protect\citeauthoryear{Kalal, Mikolajczyk, and Matas}{Kalal
  et~al\mbox{.}}{2011}]%
        {Kalal2011Tracking}
\bibfield{author}{\bibinfo{person}{Zdenek Kalal}, \bibinfo{person}{Krystian
  Mikolajczyk}, {and} \bibinfo{person}{Jiri Matas}.}
  \bibinfo{year}{2011}\natexlab{}.
\newblock \showarticletitle{Tracking-learning-detection}.
\newblock \bibinfo{journal}{\emph{IEEE Trans Pattern Anal Mach Intell}}
  \bibinfo{volume}{34}, \bibinfo{number}{7} (\bibinfo{year}{2011}),
  \bibinfo{pages}{1409--1422}.
\newblock


\bibitem[\protect\citeauthoryear{Kristan, Leonardis, Matas, Felsberg,
  Pflugfelder, \v{C}ehovin Zajc, Vojir, H\"{a}ger, Luke\v{z}i\v{c}, and
  Fernandez}{Kristan et~al\mbox{.}}{2016}]%
        {Kristan2016a}
\bibfield{author}{\bibinfo{person}{Matej Kristan}, \bibinfo{person}{Ale\v{s}
  Leonardis}, \bibinfo{person}{Jiri Matas}, \bibinfo{person}{Michael Felsberg},
  \bibinfo{person}{Roman Pflugfelder}, \bibinfo{person}{Luka \v{C}ehovin Zajc},
  \bibinfo{person}{Tomas Vojir}, \bibinfo{person}{Gustav H\"{a}ger},
  \bibinfo{person}{Alan Luke\v{z}i\v{c}}, {and} \bibinfo{person}{Gustavo
  Fernandez}.} \bibinfo{year}{2016}\natexlab{}.
\newblock \bibinfo{title}{The Visual Object Tracking VOT2016 challenge
  results}.
\newblock \bibinfo{howpublished}{Springer}.
\newblock
\urldef\tempurl%
\url{http://www.springer.com/gp/book/9783319488806}
\showURL{%
\tempurl}


\bibitem[\protect\citeauthoryear{Kristan, Leonardis, Matas, Felsberg,
  Pfugfelder, \v{C}ehovin Zajc, Vojir, Bhat, Lukezic, Eldesokey, Fernandez, and
  et~al.}{Kristan et~al\mbox{.}}{2018}]%
        {Kristan2018a}
\bibfield{author}{\bibinfo{person}{Matej Kristan}, \bibinfo{person}{Ales
  Leonardis}, \bibinfo{person}{Jiri Matas}, \bibinfo{person}{Michael Felsberg},
  \bibinfo{person}{Roman Pfugfelder}, \bibinfo{person}{Luka \v{C}ehovin Zajc},
  \bibinfo{person}{Tomas Vojir}, \bibinfo{person}{Goutam Bhat},
  \bibinfo{person}{Alan Lukezic}, \bibinfo{person}{Abdelrahman Eldesokey},
  \bibinfo{person}{Gustavo Fernandez}, {and} \bibinfo{person}{et al.}}
  \bibinfo{year}{2018}\natexlab{}.
\newblock \bibinfo{title}{The sixth Visual Object Tracking VOT2018 challenge
  results}.
\newblock
\newblock


\bibitem[\protect\citeauthoryear{Kristan, Matas, Leonardis, Felsberg,
  Pflugfelder, Kamarainen, \v{C}ehovin Zajc, Drbohlav, Lukezic, Berg,
  Eldesokey, Kapyla, and Fernandez}{Kristan et~al\mbox{.}}{2019}]%
        {Kristan2019a}
\bibfield{author}{\bibinfo{person}{Matej Kristan}, \bibinfo{person}{Jiri
  Matas}, \bibinfo{person}{Ales Leonardis}, \bibinfo{person}{Michael Felsberg},
  \bibinfo{person}{Roman Pflugfelder}, \bibinfo{person}{Joni-Kristian
  Kamarainen}, \bibinfo{person}{Luka \v{C}ehovin Zajc}, \bibinfo{person}{Ondrej
  Drbohlav}, \bibinfo{person}{Alan Lukezic}, \bibinfo{person}{Amanda Berg},
  \bibinfo{person}{Abdelrahman Eldesokey}, \bibinfo{person}{Jani Kapyla}, {and}
  \bibinfo{person}{Gustavo Fernandez}.} \bibinfo{year}{2019}\natexlab{}.
\newblock \bibinfo{title}{The Seventh Visual Object Tracking VOT2019 Challenge
  Results}.
\newblock
\newblock


\bibitem[\protect\citeauthoryear{Krizhevsky, Sutskever, and Hinton}{Krizhevsky
  et~al\mbox{.}}{2012}]%
        {krizhevsky2012imagenet}
\bibfield{author}{\bibinfo{person}{Alex Krizhevsky}, \bibinfo{person}{Ilya
  Sutskever}, {and} \bibinfo{person}{Geoffrey~E Hinton}.}
  \bibinfo{year}{2012}\natexlab{}.
\newblock \showarticletitle{Imagenet classification with deep convolutional
  neural networks}. In \bibinfo{booktitle}{\emph{Advances in neural information
  processing systems}}. \bibinfo{pages}{1097--1105}.
\newblock


\bibitem[\protect\citeauthoryear{Law and Deng}{Law and Deng}{2018}]%
        {law2018cornernet}
\bibfield{author}{\bibinfo{person}{Hei Law} {and} \bibinfo{person}{Jia Deng}.}
  \bibinfo{year}{2018}\natexlab{}.
\newblock \showarticletitle{Cornernet: Detecting objects as paired keypoints}.
  In \bibinfo{booktitle}{\emph{Proceedings of the European Conference on
  Computer Vision (ECCV)}}. \bibinfo{pages}{734--750}.
\newblock


\bibitem[\protect\citeauthoryear{Li, Wu, Wang, Zhang, Xing, and Yan}{Li
  et~al\mbox{.}}{2019b}]%
        {li2019siamrpn++}
\bibfield{author}{\bibinfo{person}{Bo Li}, \bibinfo{person}{Wei Wu},
  \bibinfo{person}{Qiang Wang}, \bibinfo{person}{Fangyi Zhang},
  \bibinfo{person}{Junliang Xing}, {and} \bibinfo{person}{Junjie Yan}.}
  \bibinfo{year}{2019}\natexlab{b}.
\newblock \showarticletitle{Siamrpn++: Evolution of siamese visual tracking
  with very deep networks}. In \bibinfo{booktitle}{\emph{Proceedings of the
  IEEE Conference on Computer Vision and Pattern Recognition}}.
  \bibinfo{pages}{4282--4291}.
\newblock


\bibitem[\protect\citeauthoryear{Li, Yan, Wu, Zhu, and Hu}{Li
  et~al\mbox{.}}{2018b}]%
        {li2018high}
\bibfield{author}{\bibinfo{person}{Bo Li}, \bibinfo{person}{Junjie Yan},
  \bibinfo{person}{Wei Wu}, \bibinfo{person}{Zheng Zhu}, {and}
  \bibinfo{person}{Xiaolin Hu}.} \bibinfo{year}{2018}\natexlab{b}.
\newblock \showarticletitle{High performance visual tracking with siamese
  region proposal network}. In \bibinfo{booktitle}{\emph{Proceedings of the
  IEEE Conference on Computer Vision and Pattern Recognition}}.
  \bibinfo{pages}{8971--8980}.
\newblock


\bibitem[\protect\citeauthoryear{Li, Tian, Zuo, Zhang, and Yang}{Li
  et~al\mbox{.}}{2018a}]%
        {li2018learning}
\bibfield{author}{\bibinfo{person}{Feng Li}, \bibinfo{person}{Cheng Tian},
  \bibinfo{person}{Wangmeng Zuo}, \bibinfo{person}{Lei Zhang}, {and}
  \bibinfo{person}{Ming-Hsuan Yang}.} \bibinfo{year}{2018}\natexlab{a}.
\newblock \showarticletitle{Learning spatial-temporal regularized correlation
  filters for visual tracking}. In \bibinfo{booktitle}{\emph{Proceedings of the
  IEEE Conference on Computer Vision and Pattern Recognition}}.
  \bibinfo{pages}{4904--4913}.
\newblock


\bibitem[\protect\citeauthoryear{Li, Ma, Wu, He, and Yang}{Li
  et~al\mbox{.}}{2019a}]%
        {li2019target}
\bibfield{author}{\bibinfo{person}{Xin Li}, \bibinfo{person}{Chao Ma},
  \bibinfo{person}{Baoyuan Wu}, \bibinfo{person}{Zhenyu He}, {and}
  \bibinfo{person}{Ming-Hsuan Yang}.} \bibinfo{year}{2019}\natexlab{a}.
\newblock \showarticletitle{Target-aware deep tracking}. In
  \bibinfo{booktitle}{\emph{Proceedings of the IEEE Conference on Computer
  Vision and Pattern Recognition}}. \bibinfo{pages}{1369--1378}.
\newblock


\bibitem[\protect\citeauthoryear{Li and Zhang}{Li and Zhang}{2019}]%
        {li2019siamvgg}
\bibfield{author}{\bibinfo{person}{Yuhong Li} {and} \bibinfo{person}{Xiaofan
  Zhang}.} \bibinfo{year}{2019}\natexlab{}.
\newblock \showarticletitle{SiamVGG: Visual tracking using deeper siamese
  networks}.
\newblock \bibinfo{journal}{\emph{arXiv preprint arXiv:1902.02804}}
  (\bibinfo{year}{2019}).
\newblock


\bibitem[\protect\citeauthoryear{Li and Zhu}{Li and Zhu}{2014}]%
        {li2014scale}
\bibfield{author}{\bibinfo{person}{Yang Li} {and} \bibinfo{person}{Jianke
  Zhu}.} \bibinfo{year}{2014}\natexlab{}.
\newblock \showarticletitle{A scale adaptive kernel correlation filter tracker
  with feature integration}. In \bibinfo{booktitle}{\emph{European conference
  on computer vision}}. Springer, \bibinfo{pages}{254--265}.
\newblock


\bibitem[\protect\citeauthoryear{Lin, Goyal, Girshick, He, and Doll{\'a}r}{Lin
  et~al\mbox{.}}{2017}]%
        {lin2017focal}
\bibfield{author}{\bibinfo{person}{Tsung-Yi Lin}, \bibinfo{person}{Priya
  Goyal}, \bibinfo{person}{Ross Girshick}, \bibinfo{person}{Kaiming He}, {and}
  \bibinfo{person}{Piotr Doll{\'a}r}.} \bibinfo{year}{2017}\natexlab{}.
\newblock \showarticletitle{Focal loss for dense object detection}. In
  \bibinfo{booktitle}{\emph{Proceedings of the IEEE international conference on
  computer vision}}. \bibinfo{pages}{2980--2988}.
\newblock


\bibitem[\protect\citeauthoryear{Lin, Maire, Belongie, Hays, Perona, Ramanan,
  Doll{\'a}r, and Zitnick}{Lin et~al\mbox{.}}{2014}]%
        {lin2014microsoft}
\bibfield{author}{\bibinfo{person}{Tsung-Yi Lin}, \bibinfo{person}{Michael
  Maire}, \bibinfo{person}{Serge Belongie}, \bibinfo{person}{James Hays},
  \bibinfo{person}{Pietro Perona}, \bibinfo{person}{Deva Ramanan},
  \bibinfo{person}{Piotr Doll{\'a}r}, {and} \bibinfo{person}{C~Lawrence
  Zitnick}.} \bibinfo{year}{2014}\natexlab{}.
\newblock \showarticletitle{Microsoft coco: Common objects in context}. In
  \bibinfo{booktitle}{\emph{European conference on computer vision}}. Springer,
  \bibinfo{pages}{740--755}.
\newblock


\bibitem[\protect\citeauthoryear{Mueller, Smith, and Ghanem}{Mueller
  et~al\mbox{.}}{2016}]%
        {mueller2016benchmark}
\bibfield{author}{\bibinfo{person}{Matthias Mueller}, \bibinfo{person}{Neil
  Smith}, {and} \bibinfo{person}{Bernard Ghanem}.}
  \bibinfo{year}{2016}\natexlab{}.
\newblock \showarticletitle{A benchmark and simulator for uav tracking}. In
  \bibinfo{booktitle}{\emph{European conference on computer vision}}. Springer,
  \bibinfo{pages}{445--461}.
\newblock


\bibitem[\protect\citeauthoryear{Nam and Han}{Nam and Han}{2016}]%
        {nam2016learning}
\bibfield{author}{\bibinfo{person}{Hyeonseob Nam} {and}
  \bibinfo{person}{Bohyung Han}.} \bibinfo{year}{2016}\natexlab{}.
\newblock \showarticletitle{Learning multi-domain convolutional neural networks
  for visual tracking}. In \bibinfo{booktitle}{\emph{Proceedings of the IEEE
  conference on computer vision and pattern recognition}}.
  \bibinfo{pages}{4293--4302}.
\newblock


\bibitem[\protect\citeauthoryear{Real, Shlens, Mazzocchi, Pan, and
  Vanhoucke}{Real et~al\mbox{.}}{2017}]%
        {real2017youtube}
\bibfield{author}{\bibinfo{person}{Esteban Real}, \bibinfo{person}{Jonathon
  Shlens}, \bibinfo{person}{Stefano Mazzocchi}, \bibinfo{person}{Xin Pan},
  {and} \bibinfo{person}{Vincent Vanhoucke}.} \bibinfo{year}{2017}\natexlab{}.
\newblock \showarticletitle{Youtube-boundingboxes: A large high-precision
  human-annotated data set for object detection in video}. In
  \bibinfo{booktitle}{\emph{Proceedings of the IEEE Conference on Computer
  Vision and Pattern Recognition}}. \bibinfo{pages}{5296--5305}.
\newblock


\bibitem[\protect\citeauthoryear{Redmon, Divvala, Girshick, and Farhadi}{Redmon
  et~al\mbox{.}}{2016}]%
        {redmon2016you}
\bibfield{author}{\bibinfo{person}{Joseph Redmon}, \bibinfo{person}{Santosh
  Divvala}, \bibinfo{person}{Ross Girshick}, {and} \bibinfo{person}{Ali
  Farhadi}.} \bibinfo{year}{2016}\natexlab{}.
\newblock \showarticletitle{You only look once: Unified, real-time object
  detection}. In \bibinfo{booktitle}{\emph{Proceedings of the IEEE conference
  on computer vision and pattern recognition}}. \bibinfo{pages}{779--788}.
\newblock


\bibitem[\protect\citeauthoryear{Ren, He, Girshick, and Sun}{Ren
  et~al\mbox{.}}{2015}]%
        {ren2015faster}
\bibfield{author}{\bibinfo{person}{Shaoqing Ren}, \bibinfo{person}{Kaiming He},
  \bibinfo{person}{Ross Girshick}, {and} \bibinfo{person}{Jian Sun}.}
  \bibinfo{year}{2015}\natexlab{}.
\newblock \showarticletitle{Faster r-cnn: Towards real-time object detection
  with region proposal networks}. In \bibinfo{booktitle}{\emph{Advances in
  neural information processing systems}}. \bibinfo{pages}{91--99}.
\newblock


\bibitem[\protect\citeauthoryear{Simonyan and Zisserman}{Simonyan and
  Zisserman}{2014}]%
        {simonyan2014very}
\bibfield{author}{\bibinfo{person}{Karen Simonyan} {and}
  \bibinfo{person}{Andrew Zisserman}.} \bibinfo{year}{2014}\natexlab{}.
\newblock \showarticletitle{Very deep convolutional networks for large-scale
  image recognition}.
\newblock \bibinfo{journal}{\emph{arXiv preprint arXiv:1409.1556}}
  (\bibinfo{year}{2014}).
\newblock


\bibitem[\protect\citeauthoryear{Szegedy, Liu, Jia, Sermanet, Reed, Anguelov,
  Erhan, Vanhoucke, and Rabinovich}{Szegedy et~al\mbox{.}}{2015}]%
        {szegedy2015going}
\bibfield{author}{\bibinfo{person}{Christian Szegedy}, \bibinfo{person}{Wei
  Liu}, \bibinfo{person}{Yangqing Jia}, \bibinfo{person}{Pierre Sermanet},
  \bibinfo{person}{Scott Reed}, \bibinfo{person}{Dragomir Anguelov},
  \bibinfo{person}{Dumitru Erhan}, \bibinfo{person}{Vincent Vanhoucke}, {and}
  \bibinfo{person}{Andrew Rabinovich}.} \bibinfo{year}{2015}\natexlab{}.
\newblock \showarticletitle{Going deeper with convolutions}. In
  \bibinfo{booktitle}{\emph{Proceedings of the IEEE conference on computer
  vision and pattern recognition}}. \bibinfo{pages}{1--9}.
\newblock


\bibitem[\protect\citeauthoryear{Tao, Gavves, and Smeulders}{Tao
  et~al\mbox{.}}{2016}]%
        {tao2016siamese}
\bibfield{author}{\bibinfo{person}{Ran Tao}, \bibinfo{person}{Efstratios
  Gavves}, {and} \bibinfo{person}{Arnold~WM Smeulders}.}
  \bibinfo{year}{2016}\natexlab{}.
\newblock \showarticletitle{Siamese instance search for tracking}. In
  \bibinfo{booktitle}{\emph{Proceedings of the IEEE conference on computer
  vision and pattern recognition}}. \bibinfo{pages}{1420--1429}.
\newblock


\bibitem[\protect\citeauthoryear{Tian, Shen, Chen, and He}{Tian
  et~al\mbox{.}}{2019}]%
        {tian2019fcos}
\bibfield{author}{\bibinfo{person}{Zhi Tian}, \bibinfo{person}{Chunhua Shen},
  \bibinfo{person}{Hao Chen}, {and} \bibinfo{person}{Tong He}.}
  \bibinfo{year}{2019}\natexlab{}.
\newblock \showarticletitle{Fcos: Fully convolutional one-stage object
  detection}. In \bibinfo{booktitle}{\emph{Proceedings of the IEEE
  International Conference on Computer Vision}}. \bibinfo{pages}{9627--9636}.
\newblock


\bibitem[\protect\citeauthoryear{Valmadre, Bertinetto, Henriques, Vedaldi, and
  Torr}{Valmadre et~al\mbox{.}}{2017}]%
        {Valmadre2017End}
\bibfield{author}{\bibinfo{person}{Jack Valmadre}, \bibinfo{person}{Luca
  Bertinetto}, \bibinfo{person}{Joao Henriques}, \bibinfo{person}{Andrea
  Vedaldi}, {and} \bibinfo{person}{Philip~HS Torr}.}
  \bibinfo{year}{2017}\natexlab{}.
\newblock \showarticletitle{End-to-end representation learning for correlation
  filter based tracking}. In \bibinfo{booktitle}{\emph{Proceedings of the IEEE
  Conference on Computer Vision and Pattern Recognition}}.
  \bibinfo{pages}{2805--2813}.
\newblock


\bibitem[\protect\citeauthoryear{Wang, Luo, Xiong, and Zeng}{Wang
  et~al\mbox{.}}{2019}]%
        {wang2019spm}
\bibfield{author}{\bibinfo{person}{Guangting Wang}, \bibinfo{person}{Chong
  Luo}, \bibinfo{person}{Zhiwei Xiong}, {and} \bibinfo{person}{Wenjun Zeng}.}
  \bibinfo{year}{2019}\natexlab{}.
\newblock \showarticletitle{Spm-tracker: Series-parallel matching for real-time
  visual object tracking}. In \bibinfo{booktitle}{\emph{Proceedings of the IEEE
  Conference on Computer Vision and Pattern Recognition}}.
  \bibinfo{pages}{3643--3652}.
\newblock


\bibitem[\protect\citeauthoryear{{Wu}, {Lim}, and {Yang}}{{Wu}
  et~al\mbox{.}}{2013}]%
        {6619156}
\bibfield{author}{\bibinfo{person}{Y. {Wu}}, \bibinfo{person}{J. {Lim}}, {and}
  \bibinfo{person}{M. {Yang}}.} \bibinfo{year}{2013}\natexlab{}.
\newblock \showarticletitle{Online Object Tracking: A Benchmark}. In
  \bibinfo{booktitle}{\emph{2013 IEEE Conference on Computer Vision and Pattern
  Recognition}}. \bibinfo{pages}{2411--2418}.
\newblock


\bibitem[\protect\citeauthoryear{{Wu}, {Lim}, and {Yang}}{{Wu}
  et~al\mbox{.}}{2015}]%
        {7001050}
\bibfield{author}{\bibinfo{person}{Y. {Wu}}, \bibinfo{person}{J. {Lim}}, {and}
  \bibinfo{person}{M. {Yang}}.} \bibinfo{year}{2015}\natexlab{}.
\newblock \showarticletitle{Object Tracking Benchmark}.
\newblock \bibinfo{journal}{\emph{IEEE Transactions on Pattern Analysis and
  Machine Intelligence}} \bibinfo{volume}{37}, \bibinfo{number}{9}
  (\bibinfo{year}{2015}), \bibinfo{pages}{1834--1848}.
\newblock


\bibitem[\protect\citeauthoryear{Xu, Feng, Wu, and Kittler}{Xu
  et~al\mbox{.}}{2019}]%
        {xu2019learning}
\bibfield{author}{\bibinfo{person}{Tianyang Xu}, \bibinfo{person}{Zhen-Hua
  Feng}, \bibinfo{person}{Xiao-Jun Wu}, {and} \bibinfo{person}{Josef Kittler}.}
  \bibinfo{year}{2019}\natexlab{}.
\newblock \showarticletitle{Learning Adaptive Discriminative Correlation
  Filters via Temporal Consistency Preserving Spatial Feature Selection for
  Robust Visual Object Tracking}.
\newblock \bibinfo{journal}{\emph{IEEE Transactions on Image Processing}}
  \bibinfo{volume}{28}, \bibinfo{number}{11} (\bibinfo{year}{2019}),
  \bibinfo{pages}{5596--5609}.
\newblock


\bibitem[\protect\citeauthoryear{Yang, Wang, Hong, Tian, and Rui}{Yang
  et~al\mbox{.}}{2017b}]%
        {yang2017enhancing}
\bibfield{author}{\bibinfo{person}{Xun Yang}, \bibinfo{person}{Meng Wang},
  \bibinfo{person}{Richang Hong}, \bibinfo{person}{Qi Tian}, {and}
  \bibinfo{person}{Yong Rui}.} \bibinfo{year}{2017}\natexlab{b}.
\newblock \showarticletitle{Enhancing person re-identification in a
  self-trained subspace}.
\newblock \bibinfo{journal}{\emph{ACM Transactions on Multimedia Computing,
  Communications, and Applications (TOMM)}} \bibinfo{volume}{13},
  \bibinfo{number}{3} (\bibinfo{year}{2017}), \bibinfo{pages}{1--23}.
\newblock


\bibitem[\protect\citeauthoryear{Yang, Wang, and Tao}{Yang
  et~al\mbox{.}}{2017a}]%
        {yang2017person}
\bibfield{author}{\bibinfo{person}{Xun Yang}, \bibinfo{person}{Meng Wang},
  {and} \bibinfo{person}{Dacheng Tao}.} \bibinfo{year}{2017}\natexlab{a}.
\newblock \showarticletitle{Person re-identification with metric learning using
  privileged information}.
\newblock \bibinfo{journal}{\emph{IEEE Transactions on Image Processing}}
  \bibinfo{volume}{27}, \bibinfo{number}{2} (\bibinfo{year}{2017}),
  \bibinfo{pages}{791--805}.
\newblock


\bibitem[\protect\citeauthoryear{Yang, Zhou, and Wang}{Yang
  et~al\mbox{.}}{2018}]%
        {yang2018person}
\bibfield{author}{\bibinfo{person}{Xun Yang}, \bibinfo{person}{Peicheng Zhou},
  {and} \bibinfo{person}{Meng Wang}.} \bibinfo{year}{2018}\natexlab{}.
\newblock \showarticletitle{Person reidentification via structural deep metric
  learning}.
\newblock \bibinfo{journal}{\emph{IEEE Transactions on Neural Networks and
  Learning Systems}} \bibinfo{volume}{30}, \bibinfo{number}{10}
  (\bibinfo{year}{2018}), \bibinfo{pages}{2987--2998}.
\newblock


\bibitem[\protect\citeauthoryear{Yu, Jiang, Wang, Cao, and Huang}{Yu
  et~al\mbox{.}}{2016}]%
        {yu2016unitbox}
\bibfield{author}{\bibinfo{person}{Jiahui Yu}, \bibinfo{person}{Yuning Jiang},
  \bibinfo{person}{Zhangyang Wang}, \bibinfo{person}{Zhimin Cao}, {and}
  \bibinfo{person}{Thomas Huang}.} \bibinfo{year}{2016}\natexlab{}.
\newblock \showarticletitle{Unitbox: An advanced object detection network}. In
  \bibinfo{booktitle}{\emph{Proceedings of the 24th ACM international
  conference on Multimedia}}. \bibinfo{pages}{516--520}.
\newblock


\bibitem[\protect\citeauthoryear{Zhang, Wang, Qi, Wang, Feng, and Lu}{Zhang
  et~al\mbox{.}}{2018}]%
        {zhang2018structured}
\bibfield{author}{\bibinfo{person}{Yunhua Zhang}, \bibinfo{person}{Lijun Wang},
  \bibinfo{person}{Jinqing Qi}, \bibinfo{person}{Dong Wang},
  \bibinfo{person}{Mengyang Feng}, {and} \bibinfo{person}{Huchuan Lu}.}
  \bibinfo{year}{2018}\natexlab{}.
\newblock \showarticletitle{Structured siamese network for real-time visual
  tracking}. In \bibinfo{booktitle}{\emph{Proceedings of the European
  conference on computer vision (ECCV)}}. \bibinfo{pages}{351--366}.
\newblock


\bibitem[\protect\citeauthoryear{Zhang and Peng}{Zhang and Peng}{2019}]%
        {zhang2019deeper}
\bibfield{author}{\bibinfo{person}{Zhipeng Zhang} {and} \bibinfo{person}{Houwen
  Peng}.} \bibinfo{year}{2019}\natexlab{}.
\newblock \showarticletitle{Deeper and wider siamese networks for real-time
  visual tracking}. In \bibinfo{booktitle}{\emph{Proceedings of the IEEE
  Conference on Computer Vision and Pattern Recognition}}.
  \bibinfo{pages}{4591--4600}.
\newblock


\bibitem[\protect\citeauthoryear{Zhu, Wang, Li, Wu, Yan, and Hu}{Zhu
  et~al\mbox{.}}{2018}]%
        {zhu2018distractor}
\bibfield{author}{\bibinfo{person}{Zheng Zhu}, \bibinfo{person}{Qiang Wang},
  \bibinfo{person}{Bo Li}, \bibinfo{person}{Wei Wu}, \bibinfo{person}{Junjie
  Yan}, {and} \bibinfo{person}{Weiming Hu}.} \bibinfo{year}{2018}\natexlab{}.
\newblock \showarticletitle{Distractor-aware siamese networks for visual object
  tracking}. In \bibinfo{booktitle}{\emph{Proceedings of the European
  Conference on Computer Vision (ECCV)}}. \bibinfo{pages}{101--117}.
\newblock


\end{thebibliography}

\end{document}